\documentclass[journal]{IEEEtran}
\usepackage{amsmath,amsfonts}
\usepackage{algorithm,algpseudocode}
\usepackage{array}
\usepackage{multirow}
\usepackage{caption}
\usepackage{subcaption}
\usepackage{graphicx}
\usepackage{textcomp}
\usepackage{stfloats}
\usepackage{url}
\usepackage{verbatim}
\usepackage{graphicx}
\usepackage{cite}
\usepackage{booktabs}
\usepackage{bm}
\usepackage{color}
\usepackage{hyperref}
\usepackage{enumitem}
\usepackage{siunitx}
\usepackage{makecell}

\begin{document}
\title{Adversarial Safety-Critical Scenario Generation using Naturalistic Human Driving Priors}
\author{Kunkun~Hao, Wen~Cui,~\IEEEmembership{Member,~IEEE}, Yonggang~Luo, Lecheng~Xie, Yuqiao~Bai, Jucheng~Yang, Songyang~Yan,  Yuxi~Pan,~\IEEEmembership{Member,~IEEE}, Zijiang~Yang,~\IEEEmembership{Senior Member,~IEEE}
\thanks{Manuscript received October 30, 2023; accepted
November 14, 2023. This work was supported by the National Natural Science Foundation of China under Grants No.62232008 and No. 62032010, the National Key R\&D program of China No.2022YFB4300700, and also by Qin Chuangyuan cited the high-level innovative and entrepreneurial talent project QCYRCXM-2022-21. (\em Corresponding author: Zijiang Yang.)}
\thanks{Kunkun Hao, Yuqiao Bai, and Yuxi Pan are with the Research Center of Synkrotron, Inc., Xi’an, 710075,  China. Email: \href{haokunkun@synkrotron.ai}{haokunkun@synkrotron.ai}, \href{baiyuqiao@synkrotron.ai}{baiyuqiao@synkrotron.ai}, and \href{yuxip@synkrotron.ai}{yuxip@synkrotron.ai}.}
\thanks{Wen Cui is with the Institute for Interdisciplinary Information Core Technology, Xi’an, 710075, China, and with the Academy of Advanced Interdisciplinary Research, Xidian University, Xi’an, 710071, China, and also with the Research Center of Synkrotron, Inc., Xi’an, 710075, China. Email: \href{cuiw@iiisct.com}{cuiw@iiisct.com}.}
\thanks{Yonggang Luo, Lecheng Xie, and Jucheng Yang are with the AI Laboratory, Chongqing Changan Automobile Co. Ltd, Chongqing, 400023, China. Email: \href{luoyg3@changan.com.cn}{luoyg3@changan.com.cn}, \href{xielc@changan.com.cn}{xielc@changan.com.cn}, \href{yangjc3@changan.com.cn}{yangjc3@changan.com.cn}.}
\thanks{Songyang Yan and Zijiang Yang are with the Department of Computer Science and Technology, Xi'an Jiaotong University, Xi'an, 710049, China, and also with Synkrotron, Inc., Xi’an, 710075, China. Email: \href{tayyin@stu.xjtu.edu.cn}{tayyin@stu.xjtu.edu.cn}, \href{yang@synkrotron.ai}{yang@synkrotron.ai}.

Digital Object Identifier 10.1109/TIV.2023.3335862}
}

\maketitle

\begin{abstract}

Evaluating the decision-making system is indispensable in developing autonomous vehicles, while realistic and challenging safety-critical test scenarios play a crucial role. Obtaining these scenarios is non-trivial, thanks to the long-tailed distribution, sparsity, and rarity in real-world data sets. To tackle this problem, in this paper, we introduce a natural adversarial scenario generation solution using naturalistic human driving priors and reinforcement learning techniques. By doing this, we can obtain large-scale test scenarios that are both diverse and realistic. Specifically, we build a simulation environment that mimics natural traffic interaction scenarios. Informed by this environment, we implement a two-stage procedure. The first stage incorporates conventional rule-based models, e.g., IDM~(Intelligent Driver Model) and MOBIL~(Minimizing Overall Braking Induced by Lane changes) model, to coarsely and discretely capture and calibrate key control parameters from the real-world dataset. Next, we leverage GAIL~(Generative Adversarial Imitation Learning) to represent driver behaviors continuously. The derived GAIL can be further used to design a PPO~(Proximal Policy Optimization)-based actor-critic network framework to fine-tune the reward function, and then optimize our natural adversarial scenario generation solution. Extensive experiments have been conducted in two popular datasets, NGSIM and INTERACTION. Essential traffic parameters were measured in comparison with the baseline model, e.g., the collision rate, accelerations, steering, and the number of lane changes. Our findings demonstrate that the proposed model can generate realistic safety-critical test scenarios covering both naturalness and adversariality with an advanced 44\% efficiency gain over the baseline model, which can be a cornerstone for the development of autonomous vehicles.

\end{abstract}


\section{Introduction}

Autonomous driving that aims to significantly improve traffic efficiency and reduce traffic accidents is empowered by recent advances in Internet-of-Things, Artificial Intelligence, and V2X at an unprecedented pace~\cite{wang2021china,ioana2022automotive,lu2014connected,cui2020sigmix,wang2020parallel,muhammad2020deep,mao2021internet,cao2022future,chen2022milestones}. To truly realize autonomous driving in practice, however, still requires enormous work due to the complex nature of autonomous driving systems and the real-world long-tail distributed scenarios~\cite{zhao2021comparative,li2022features,du2022autonomous,han2018parallel,li2022novel,li2016intelligence}. According to a recent survey from the National Highway Traffic Safety Administration~(NHTSA) in the US~\cite{summary2022report1,summary2022report2}, 522 accidents were reported related to Autonomous Vehicles~(AVs). In particular, the pioneer companies, such as Tesla and Waymo, account for the majority of these accidents, i.e., \qty{70}{\percent} of the Level 2 system from Tesla and \qty{50}{\percent} of the Level 3 to 5 systems from Waymo. Given these painful results, it becomes commonly accepted that a large number of testings and validations have to be conducted for AVs, especially for extreme scenarios, before the real world uses~\cite{li2020theoretical,ma2022verification,li2019parallel, AAS-CN-2022-0820,kalra2016driving}.

\begin{figure}
    \centering
    \begin{subfigure}{.49\linewidth}
        \centering
    	\includegraphics[width=\linewidth]{{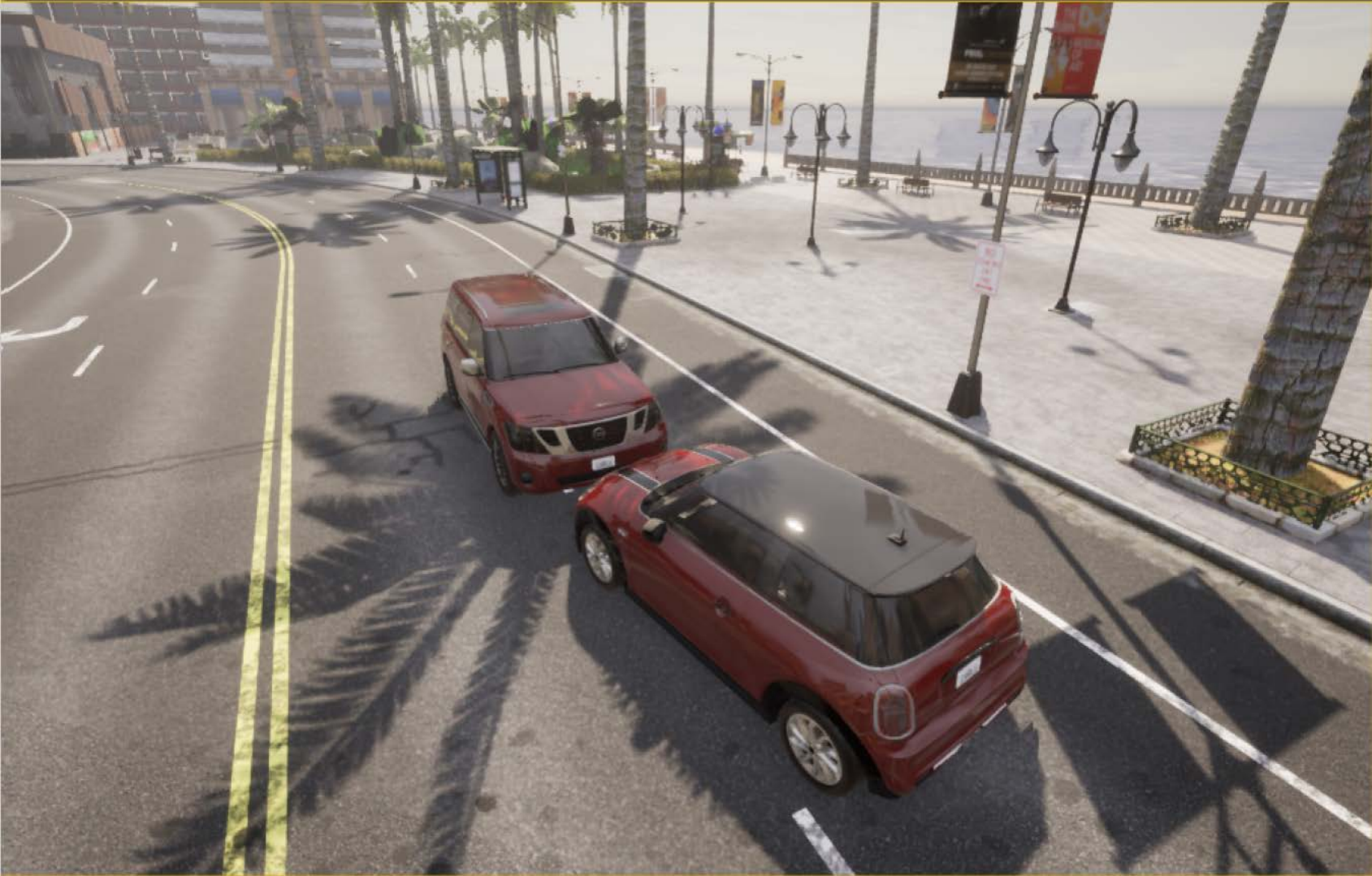}}
        \caption{Front-to-front.}
    \end{subfigure}
    \begin{subfigure}{.49\linewidth}
        \centering
    	\includegraphics[width=\linewidth]{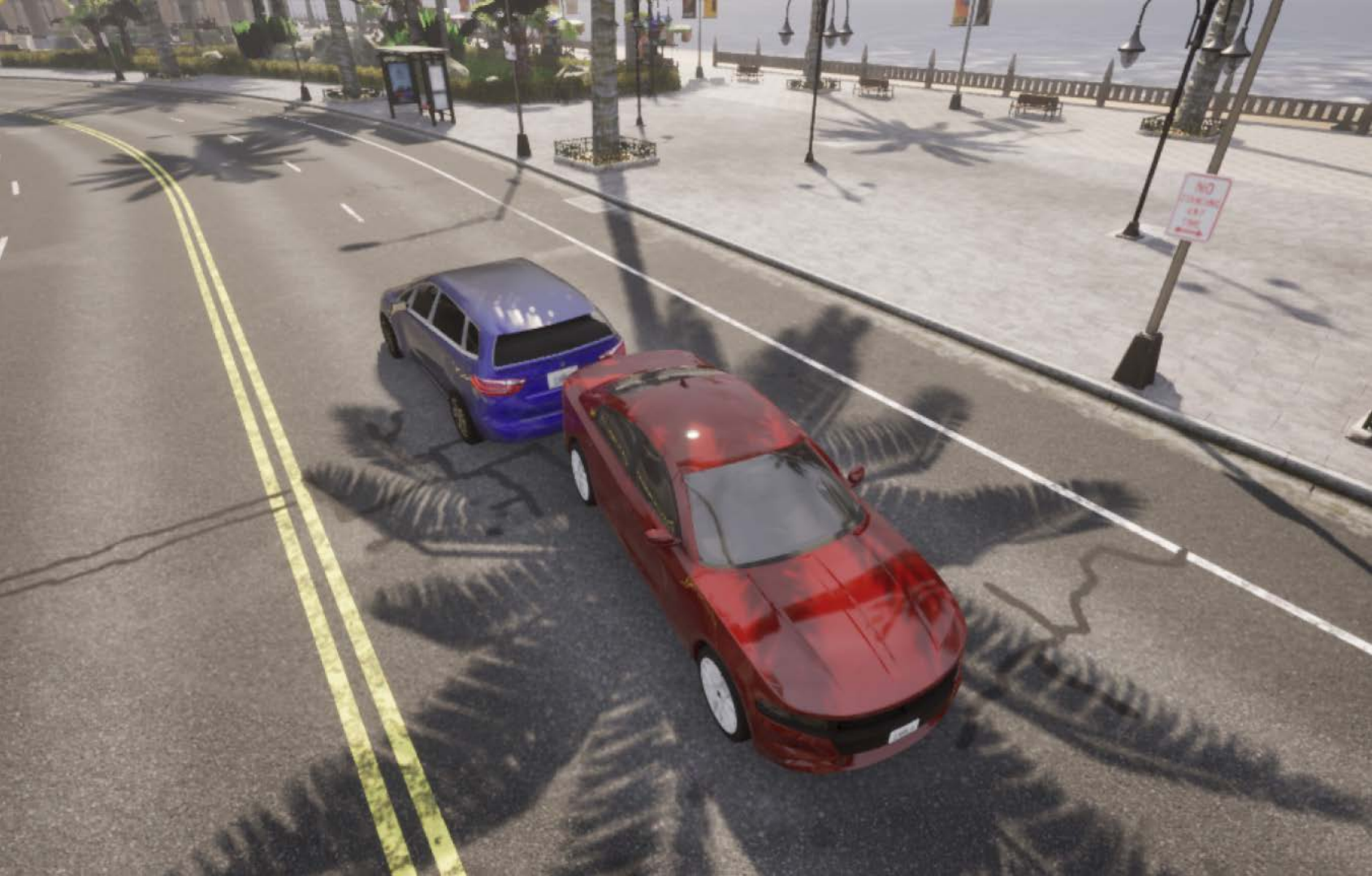}
        \caption{Rear-to-rear.}
    \end{subfigure}
    \begin{subfigure}{.49\linewidth}
        \centering
    	\includegraphics[width=\linewidth]{{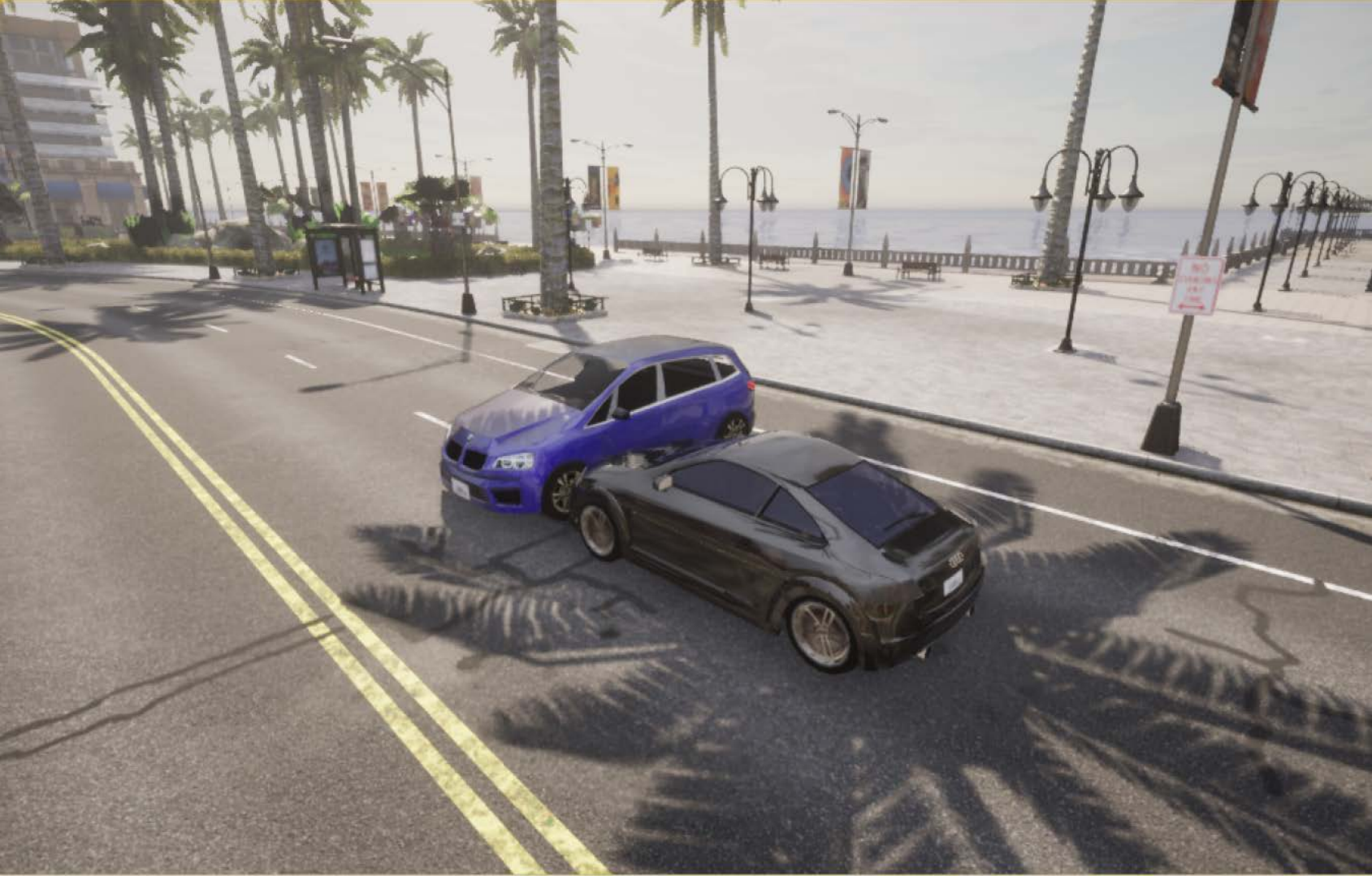}}
        \caption{T-shaped-front.}
    \end{subfigure}
    \begin{subfigure}{.49\linewidth}
        \centering
    	\includegraphics[width=\linewidth]{{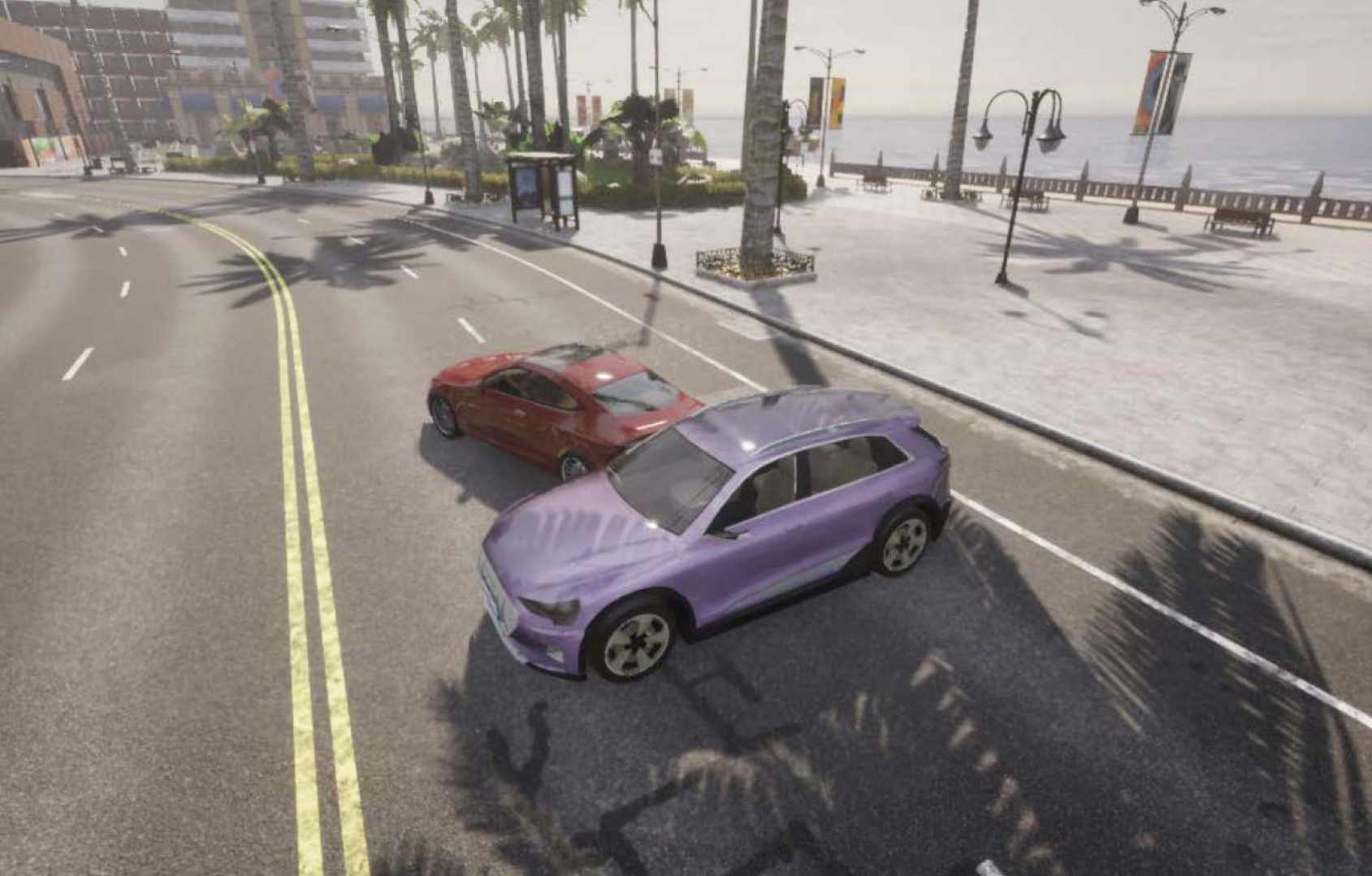}}
        \caption{T-shaped-rear.}
    \end{subfigure}
    \caption{Examples of counter-intuitive adversarial behaviors in an urban scenario. @CARLA~\cite{dosovitskiy2017carla}}
    \label{fig_1}
\end{figure}

To generate safety-critical scenarios without manual changes, existing work designed their solution according to adversarial metrics solely~\cite{norden2019efficient,feng2021intelligent, tian2023vistagpt,song2023identifying,klischat2019generating,wang2021advsim,bogdoll2023one}, e.g., collision rates, hazard distances, etc. We observe that naturalness plays an equally important role as adversariality metrics. In broad terms, naturalness and adversariality are already two critical characteristics of general generation solutions~\cite{zhang2022adversarial,wang2021human}, and we highlight their unique importance in the context of safety-critical scenario generation for autonomous vehicles~\cite{wachi2019failure}. Formally, {\em adversariality}, refers to complex driving challenges presented to AVs. An adversary scenario, including sudden lane changes, unseen jaywalkers, or dangerous maneuvering in heavy traffic, pushes the limit of driving systems and is thus crucial for the robustness verification of AVs. Furthermore, adversary scenarios are also rare in the real world, making scenario generation of this kind particularly of interest to AV designers. {\em Naturalness}, on the other hand, refers to how a generated scenario reflects real-world conditions. Obeying vehicle dynamics~\cite{rajamani2011vehicle}, understanding traffic rules, and more importantly, smoothly interacting with heterogeneous traffic participants are all defining features of the naturalness of a generated scenario. Therefore, naturalness determines whether or not scenarios are applicable in the validation of AVs. In summary, both adversariality and naturalness are crucial for scenario generation solutions, in making scenarios not only challenging to the tested AVs but also representative of real-world driving. Consequentially, after ignoring the naturalness, existing solutions may generate many counter-intuitive scenarios, e.g., front-to-front collisions, rear-to-rear collisions, and T-shape collisions, as shown in Fig.~\ref{fig_1}, resulting in ineffective tests and misleading safety reports. 

In contrast to existing work, in this paper, we present a natural adversarial scenario generation solution that considers adversariality and naturalness simultaneously. In a nutshell, we start by building a natural traffic simulation training environment, including calibrated parameters of driver models from real-world traffic datasets, i.e. NGSIM~\cite{leurent2018environment} and INTERACTION~\cite{zhan2019interaction}. In that environment, the dynamic interaction and highly uncertain driving behaviors can be represented in great detail. Based on that, we construct a human driving prior model by using generative adversarial imitation learning, and we use this model as a natural adversarial reward function to measure the naturalness of actions, which eventually enables the design of our reinforcement learning-based natural adversarial scenario generation solution. We have conducted extensive experiments using popular real-world datasets, and the results reveal that our solution generates more realistic adversarial behaviors when compared with baseline models, by achieving both adversariality and naturalness.

The contributions of this work have been concluded as follows. 
\begin{enumerate}
\item{To the best of our knowledge, we are the first to use naturalistic human driving priors and reinforcement learning to propose a large-scale and realistic safety-critical scenario generation solution. The proposed solution can be a cornerstone in testing the decision-making systems of AVs.}
\item{To pursue the realistic feature in generating adversarial scenarios, we introduce a novel reward function for the reinforcement learning agent, in which the adversariality and naturalness of adversarial behaviors are included simultaneously.}
\item{We demonstrate the performance of the proposed solution based on a real-world dataset, including the statistical analysis of the lane change rates and collision rates of vehicles, the distribution of decision-making actions, and the visual clustering of collision types on 2000 generated test scenarios.}
\end{enumerate}

We present the structure of this paper as follows. In Section~\ref{related work}, we present a comprehensive review of existing work and highlight the corresponding comparisons. In Section~\ref{design}, we elaborate on the design details including the modeling and algorithms design and the network structure. The experiment results are presented in Section~\ref{experiment}. We conclude our work and discuss potential future directions in~\ref{conculsion}.

\section{Related works}
\label{related work}
Adversarial scenario generation solutions are mainly focused on the testing and validation of perception, decision-making, and planning of AVs, and in this section, we elaborate on these advances separately. 

\subsection{Adversarial Scenario Generation for Perception}

Perception systems in AVs are empowered by the technological leap in deep learning~\cite{du2023chat,gao2023chat}. Major applications, such as object classification, object detection, and scene segmentation, rely on different types of Deep Neural Networks~(DNNs)~\cite{krizhevsky2017imagenet,he2016deep,xu2019ternary,chen2021engineering,wei2021graph}. However, these DNNs have recently been proven vulnerable to adversarial attacks~\cite{goodfellow2014explaining,yuan2019adversarial,li2023novel,buckner2020understanding,chen2022interpretive}. In particular, a small malicious modification of the perceived images, e.g., slight value changes of the partial pixels, may cause DNN's output to be significantly different, leading to misclassification or false detection.

To improve the robustness of perception systems, adversarial scenario generation solutions become indispensable. Two classical attacks of the adversarial scenario have been studied in this area, i.e., white-box attacks and black-box attacks, depending on whether the attacker knows the model details or not. For white-box attacks, Hendrik~\textit{et al}.~\cite{hendrik2017universal} designed adversarial perturbations using gradient-dependent optimization, and then used them to attack against the street view image semantic segmentation. Other than that, robust physical perturbations were introduced by Eykholt~\textit{et al}. for physical-world objects~\cite{eykholt2018robust}. By doing this, an example showed that a stop sign classifier can be consistently misled by the physical perturbations, even under different physical settings. For black-box attacks, Xiong~\textit{et al}.~\cite{xiong2021multi} proposed multi-source adversarial sample attack models for the attacking of image and LiDAR perception systems together, including a parallel attack model and a fusion attack model to cover different sources of the perceived information. Morgulis~\textit{et al}.~\cite{morgulis2019fooling} provided an adversarial traffic sign generator that fooled a range of production-grade traffic sign recognition systems, and demonstrated the effectiveness of the generator through real-world experiments. Furthermore, Li~\textit{et al}.~\cite{li2020adaptive} presented a perturbation generator for traffic sign images with efficient sampling, putting great risks to AVs with a high-frequency attacker. Overall, adversarial scenario generation for perception belongs to frame-level adversarial testing, making it inadequate for decision-making and planning that require continuously interact with dynamic, uncontrollable, and diverse surrounding environments, i.e., scenario-level testing. 

\subsection{Adversarial Scenario Generation for Decision-Making and Planning}

Decision-making and planning are much more complex when compared with perception solely, given the dynamically varying neighbor vehicles, the environments, and the movement of the ego vehicle itself. Henceforth, extra efforts in generating adversarial scenarios are needed, meanwhile, great attention has been drawn to this area to tackle the problem~\cite{norden2019efficient,klischat2019generating,guo2023vectorized,wang2021advsim,bogdoll2023one,ding2021multimodal,chen2021adversarial,zhang2020robust}. Straightforwardly, adversarial scenarios can come from probabilistic models using real-world traffic dataset~\cite{feng2021intelligent}. Unlike DNNs, the probabilistic model may lose certain details of the data, making them scenario-depended and hard to be extended. Authors in~\cite{ding2020learning} proposed to use the probabilistic joint distribution from different scenes to design adversarial reward functions, and then feed it into reinforcement learning models for safety-critical scenario generation. The follow-up work~\cite{ding2021multimodal} introduced adaptive sampling strategies to improve the sampling efficiency of the multidimensional scene parameter, and then multimodal safety-critical test scenarios can be covered. These mentioned solutions leveraged optimized initial parameters for each scene, and therefore can quickly generate straightforward scenarios but limited their ability to reflect real-world driving behaviors after the initialization. Recent work attempted to use trajectory priors to generate safety-critical scenarios for path planning~\cite{rempe2022generating,hanselmann2022king}. In doing this, it can get a clear view of the whole driving period generally, but the driver's behavior has not been fully learned, weakening their ability to cover decision-making adversarial scenarios that require more detailed driving information. Inspired by this work and aiming for both decision-making and planning scenarios, we develop a learning model based on generative adversarial imitation learning. This learning model can closely mimic drivers' behaviors and then can truly release the potential of human driving priors for scenario generation. The closest work to ours utilized detailed real-time traffic interactions~\cite{chen2021adversarial,sun2021corner,schott2022improving}, however, without concerning the naturalness, they turn all their attention to the adversariality of the generated scenarios. The lack of naturalness may lead to many counter-intuitive scenarios as discussed in Introduction, while we concentrate on adversariality and naturalness simultaneously.
\begin{figure*}
\centering
\includegraphics[width=6.5in]{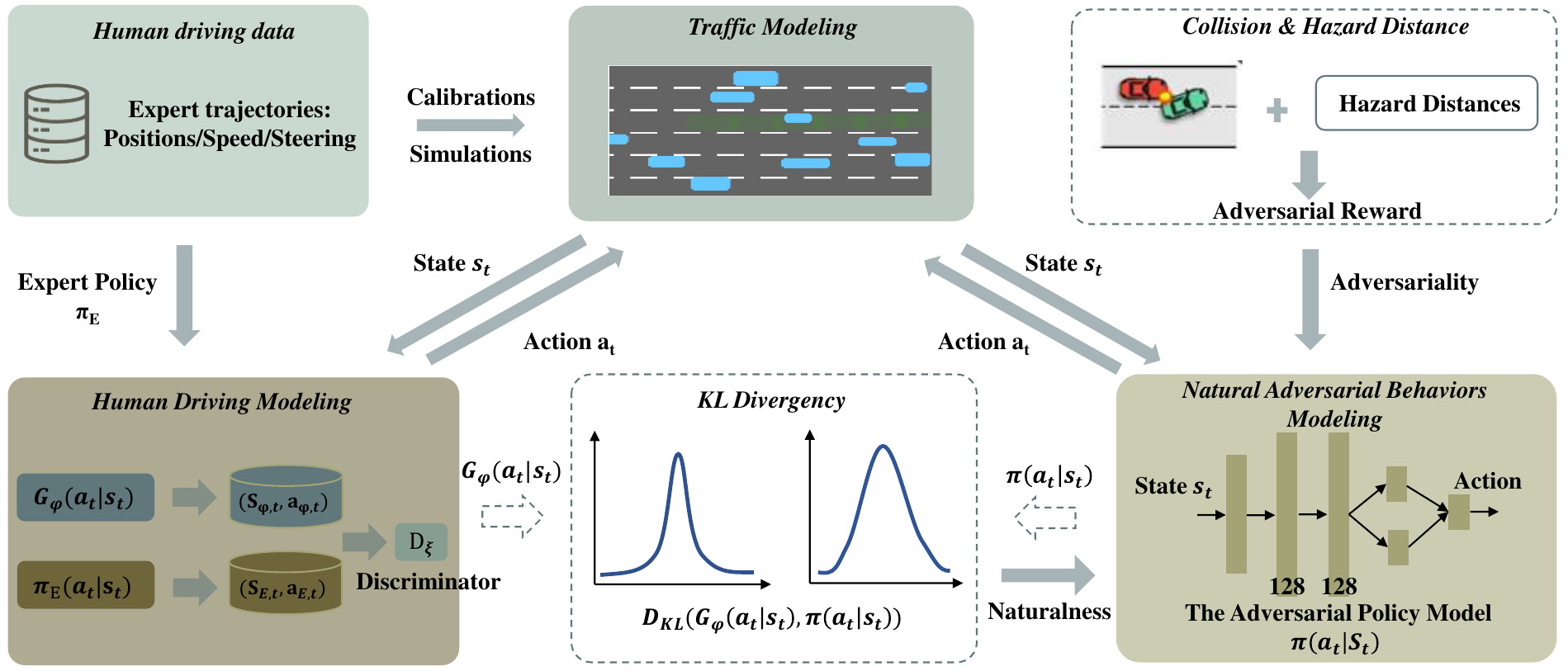}%
\caption{The overall framework of our safety-critical scenario generation solution.}
\label{fig_2}
\end{figure*}

\section{Safety-critical scenario generation framework}
\label{design}

Using a human driving dataset and reinforcement learning, we construct the overall framework of our safety-critical scenario generation solution as shown in Fig.~\ref{fig_2}, including a traffic simulation training environment, a human driving prior model, and a natural adversarial policy model.

The traffic simulation training environment includes a real-world dataset and a simulation platform that underpins our models. Without loss of generality, we use two popular real-world datasets, NGSIM~\cite{leurent2018environment} for highway scenarios and INTERACTION~\cite{zhan2019interaction} for urban scenarios, and the Highway-env simulation platform~\cite{highway-env} for their quality of data and accessibility. The data structure of the INTERACTION dataset differs from that of NGSIM. To eliminate the difference, we supplement the missing items in the INTERACTION dataset through straightforward calculations, such as determining the ID of the lane and the front car. Furthermore, it is safe to assume the AVs in the traffic are capable of perceiving their surrounding environments with certain accuracy~\cite{kar2019meta}, and we see AVs as black boxes.

Based on the above-mentioned environment, we then build our human driving prior model and natural adversarial policy model. To put it simply, we first use the traditional rule-based models, i.e., the Intelligent Driver Model~(IDM)~\cite{treiber2000congested} and the MOBIL~\cite{treiber2016mobil} model, to coarsely capture and calibrate the key parameters of longitudinal~(related to IDM) and lateral~(related to MOBIL) controls from the real-world dataset. The two mentioned models lay the foundation of a driving behavior model, but they portray behaviors in a rigid and discrete fashion, resulting in a less realistic depiction of actual driving behavior compared with learning-based solutions. Henceforth, inspired by recent advances in solving AV-related problems using reinforcement learning, such as single-agent decision-making~\cite{zhu2020safe}, multi-agent interactions~\cite{zhou2022multi}, and robot controls~\cite{johannink2019residual}, we build our human driving prior model by using Generative Adversarial Imitation Learning~(GAIL) in which continuous acceleration and steering actions can be fully represented. Moreover, this model enables the naturalness supervision of our natural adversarial policy model, and by further using the Proximal Policy Optimization~(PPO)~\cite{schulman2017proximal} in the actor-critic network framework~\cite{konda1999actor}, we complete our natural adversarial policy model. The details of our design have been provided in the following subsections.

\subsection{The Usage of IDM and MOBIL Models}

\subsubsection{Preliminaries of IDM}

The IDM is a parametric car-following model, i.e., longitudinal, in which acceleration is calculated as

\begin{equation} \label{eq_17}
    \dot{v}_{t+1}=a_{_{\text{IDM}}} \left[ 1-(\frac{v_t}{v_{_\text{IDM}}})^{\delta_{_\text{IDM}}}-(\frac{s_{_\text{IDM}}(v_t,\Delta v_t)}{s_t})^2\right]
\end{equation}

where $a_{_\text{IDM}}$ is the maximum acceleration of the vehicle, $v_t$ and $v_{_\text{IDM}}$ the current and desired speed of the vehicle, respectively, $\delta_{_\text{IDM}}$ the acceleration exponent, $\Delta v_t$ the speed difference between the vehicle and its front vehicle, $s_t$ and $s_{_\text{IDM}}$ the relative distance and the desired following distance between the vehicle and the front vehicle, respectively. We can get $s_{_\text{IDM}}$ from
\begin{equation} \label{eq_18}
    s_{_\text{IDM}}(v_t,\Delta v_t)=s_0+v_tT+\frac{v_t\Delta v_t}{2\sqrt{a_{_\text{IDM}}b_{_\text{IDM}}}}
\end{equation}
where $s_0$ is the minimum vehicle distance, $T$ the safe headway time distance, and $b_{_\text{IDM}}$ the comfortable deceleration value.

\subsubsection{Data Preprocessing of IDM} The IDM model starts with extracting car-following scenarios from the real-world trajectories. In alignment with exiting work~\cite{kurtc2016calibrating,lu2009fundamental,yu2019modified}, we first extract car-following scenarios by choosing different lead vehicles, and next, we conduct a screening for each scenario to filter out the irrelevant and erroneous scenarios. Finally, we correct and then smooth the data before the use of parameter calibration. Detailed procedures of this process are provided in Algo.~\ref{alg:IDM}, and the results are presented in Sec.~\ref{experiment}.

\begin{algorithm}[t]
\caption{Data Preprocessing of IDM}
\label{alg:IDM}
\begin{enumerate}

\item \textbf{Car-following scenarios pre-extraction:} By choosing different lead vehicles, we extract the car-following scenario from every trajectory.

\item \textbf{Scenario screening:} The irrelevant and erroneous scenarios will be excluded from subsequent stages of modeling.
\begin{enumerate}[label=\arabic*., leftmargin=*] 
\item Exclude lane-change scenarios.
\item {\bf NGSIM: }Exclude scenarios that last shorter than 30s, i.e., 300 frames at a 10Hz sampling frequency.
\item {\bf INTERACTION: }Exclude scenarios that last shorter than 4s, i.e., 40 frames at a 10Hz sampling frequency.
\item Exclude scenarios in which the ego vehicle travels less than 20m.
\item Exclude the first and last 5 seconds of each scenario clip.
\item {\bf NGSIM: }Exclude scenarios where the ego car is in the far-right lane or ramp.
\item Exclude scenarios where the instantaneous acceleration is greater than 8$\unit{m/s^2}$.
\item Exclude scenarios not for small cars (i.e., The Vehicle Class label is not equal to 2).
\item Exclude scenarios where the distance to the front car is less than 0.1m.
\end{enumerate}

\item \textbf{Data correction and smoothing:} Due to the nature of noisy data, further correction and smoothing are required.
\begin{enumerate}[label=\arabic*., leftmargin=*] 
\item Smooth the trajectory data by using the sEMA filter~\cite{thiemann2008estimating}.
\item Correct the speed and acceleration using continuous coordinates of vehicles.
\item Correct the following distance according to the length of the front vehicle.
\end{enumerate}

\end{enumerate}
\end{algorithm}

\subsubsection{Parameter Calibration of IDM} According to Eq.~\ref{eq_17} and Eq.~\ref{eq_18}, the acceleration process makes the parameter calibration of IDM a non-linear optimization problem. Promising tools can readily be adopted in solving this problem, e.g., a genetic optimizer in the MATLAB optimization solver~\cite{saraswat2013genetic,mirjalili2019evolutionary,duong2010calibration,kesting2008calibrating,yu2019modified}, while the objective function plays a key role. This objective function is dedicated to minimizing the difference between the model-generated data and the real-world dataset, and we use the relative distance~\cite{kesting2008calibrating} to represent that difference. Additionally, to reduce the bias of error measures, we leverage a mixed error measure~\cite{kurtc2016calibrating,yu2019modified} as the objective function. In particular, Eq.~\ref{eq_idm} describes the objective function $\mathcal{F}[d_\text{Sim}]$ as follows.

\begin{equation} \label{eq_idm}
    \mathcal{F}[d_\text{Sim}] = \sqrt{\frac{1}{\langle |d_\text{Data}| \rangle}\biggl< \frac{(d_\text{Data} - d_\text{Sim})^2}{|d_\text{Data}|}\biggr>},
\end{equation}

where $d_\text{Sim}$ denotes the simulated data, i.e., the relative distance generated from IDM, $d_\text{Data}$ comes from the dataset, and $\langle \cdot \rangle$ refers to the averaging of a time series data.

\subsubsection{Preliminaries of MOBIL}

The MOBIL model determines whether a vehicle should change lanes or not, i.e., lateral. It relies on two factors: one is the increase in acceleration that a vehicle would experience after a lane change, and the other is the braking variations affected by the lane change~\cite{treiber2009modeling}. In particular, we can use two specific criteria to describe the rules of MOBIL, i.e., the safety criterion and the incentive criterion. The safety criterion, as shown in Eq.~\ref{eq:safety}, defines that the deceleration of the new follower $\tilde{a}_n$ after a lane change should not be bigger than $\text{MAX\_BRAKING\_IMPOSED}$.

\begin{equation}
    \label{eq:safety}
    \tilde{a}_n \geq-\text{MAX\_BRAKING\_IMPOSED}
\end{equation}

The incentive criterion, as shown in Eq.~\ref{eq:incentive}, defines the acceleration gain from this lane change should not be smaller than $\Delta a_{\mathrm{th}}$, where $p$ is a politeness factor ranged from 0 to 1. 

\begin{equation}
    \label{eq:incentive}
    \underbrace{\tilde{a}_c-a_c}_{\text {driver }}+p(\underbrace{\tilde{a}_n-a_n}_{\text {new follower }}+\underbrace{\tilde{a}_o-a_o}_{\text {old follower }})>\Delta a_{\mathrm{th}}
\end{equation}

In Eq.~\ref{eq:incentive}, \(\tilde{a}_c\), \(a_c\), \(\tilde{a}_n\), \(a_n\), \(\tilde{a}_o\), and \(a_o\) represent specific accelerations related to the driver, new follower, and old follower, while \(\Delta a_{\mathrm{th}}\) is a threshold value.

\subsubsection{Data Preprocessing of MOBIL} The data preprocessing is similar to IDM in Algo.~\ref{alg:IDM}, except that MOBIL retains all lane-change scenarios.

\subsubsection{Parameter Calibration of MOBIL} In contrast to the non-linear optimization problem in IDM, it is straightforward that the parameters of MOBIL—specifically, $\text{MAX\_BRAKING\_IMPOSED}$ and $\Delta a_{\mathrm{th}}$—can be derived using statistical analysis of the trajectory dataset. Note that different datasets may contain bias on the parameters, and therefore, the parameters of MOBIL are often defined empirically~\cite{huang2021driving,wang2022evaluation,kesting2007general}. We illustrate the results and analysis in Sec.~\ref{experiment}. 

\subsection{Problem Formulation for Reinforcement Learning}

To begin with, we introduce a solution for completing a reinforcement learning task generally. The reinforcement learning task is usually described by the Markov Decision Process~(MDP) which assumes the next state is only related to its current state and action. Mathematically, we can represent the process as $\mathcal{M}=(\mathcal{S}, \mathcal{A}, P(s_{t+1}|s_{t}, a_{t}), r, \gamma)$, where $\mathcal{S}$ denotes the state space observed by the reinforcement learning model from its input, $\mathcal{A}$ the action space to the output of the model, representing the behavioral decision of AVs in our case. $P(s_{t+1}|s_{t}, a_{t})$ is the state transfer probability distribution, describing the dynamics of environments and representing the probability of turning to the next state $s_{t+1}$ in taking an action $a_{t}$ by an agent at state $s_{t}$, $r$ the reward function of the immediate reward to an agent for switching from state $s_{t}$ to state $s_{t+1}$, and $\gamma\in(0, 1]$ the discount factor.

The goal of a reinforcement learning task is to output a policy that maximizes the cumulative long-term discounted reward, i.e.,

\begin{equation} \label{eq_1}
    \max\limits_{\pi} \underset{\tau\sim P_\pi(\tau)}{\mathbb{E}}[\sum\limits_{t=0}^T\gamma^tr(s_t,a_t)]
\end{equation}
where $\tau$ represents the driving trajectory of the agent, $P_\pi(\tau)$ the trajectory distribution based on the policy $\pi$, and $T$ the step length of the environment.

In solving the above objective function, we can estimate the state or action-value function, which is defined recursively by the state and action-value functions based on the Bellman Optimal formula, i.e., 
\begin{equation} \label{eq_2}
    V^\pi(s_t)= \underset{a_t\sim\pi(\cdot|s_t)}{\mathbb{E}}[Q^\pi(s_t,a_t)]
\end{equation}

\begin{equation} \label{eq_3}
    Q^\pi(s_t,a_t)=r(s_t,a_t)+\gamma\underset{s_{t+1}\sim T(\cdot|s_t,a_t)}{\mathbb{E}}[V^\pi(s_{t+1})].
\end{equation}

For Equations (\ref{eq_2}) and (\ref{eq_3}), once the function $V^{\pi^\ast}(s)$ that maximizes the state value of all states $s\in S$ is found, the corresponding policy $\pi^\ast$ would be the target solution for the reinforcement learning task.




\subsection{The Proximal Policy Optimization Algorithm}

\begin{figure}
\centering
\includegraphics[width=\linewidth]{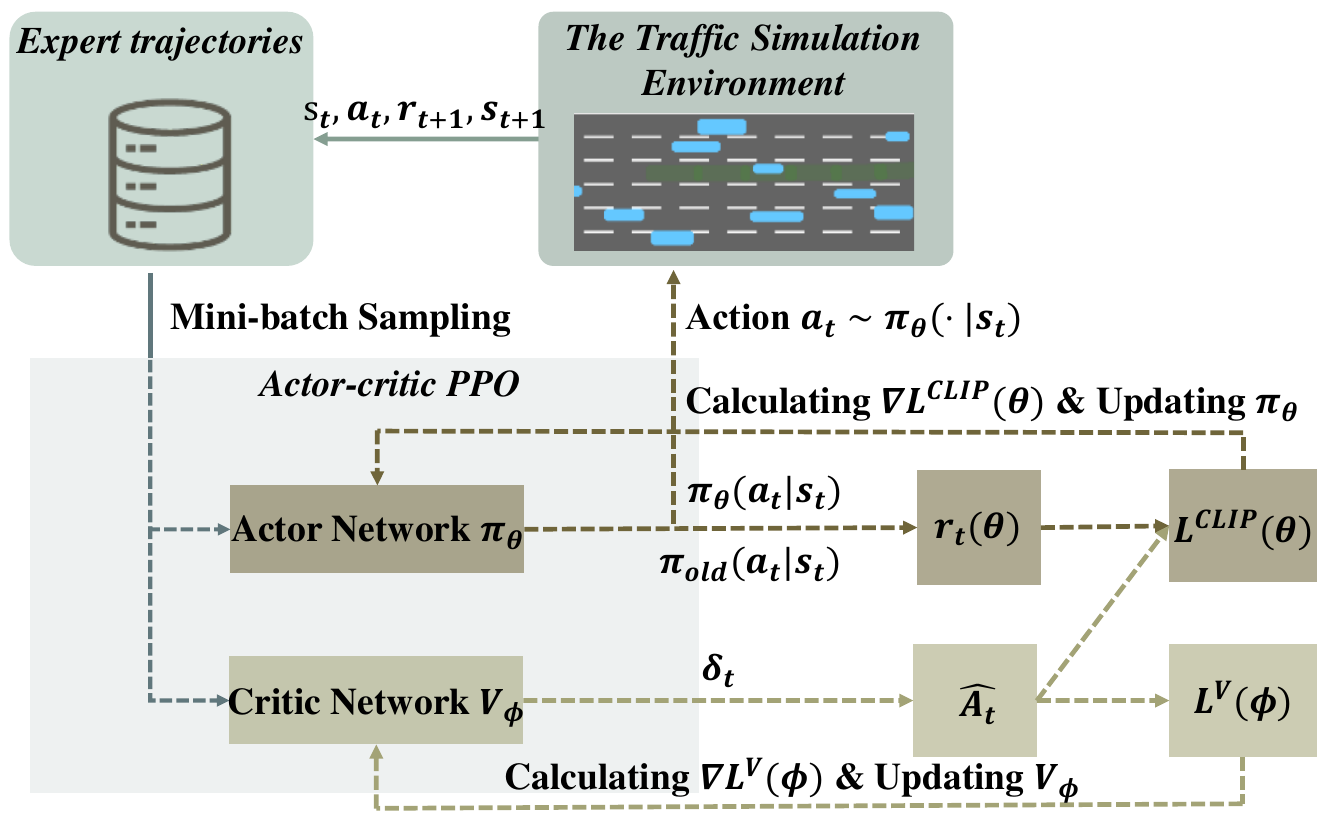}
\caption{The update process of actor-network and critic-network of PPO algorithms.}
\label{fig_3}
\end{figure}

PPO algorithms are deep reinforcement learning algorithms based on the actor-critic framework. They represent online policies and are known as model-free, aiming for the generation of both discrete and continuous controls. By using PPO, problems in continuous action space, specifically those related to autonomous driving decision tasks, can be well-addressed. These algorithms consist of two main components, an actor network that relies on the policy gradient, and a critic network that relies on the value evaluation. In this structure, the actor network generates continuous actions in response to the observed information from environments, while the critic network evaluates the output action of the actor network, influencing the subsequent action selection of the actor. PPO algorithms outperform traditional general policy gradient algorithms by achieving stable convergence speeds. In a general policy gradient algorithm, if the update step size is too large, the algorithm may easily diverge; conversely, if the update step size is too small, it may lead to convergence, but at an unacceptable speed. The procedures of PPO algorithms include running a traffic simulator through the traffic simulation environment to form a trajectories database. By obtaining different trajectories, i.e., $(s_t,a_t,r_{t+1},s_{t+1})$, small batch sampling of the trajectory database can be performed to optimize the training process of both the actor network and critic network. We provide the training update details in Fig.~\ref{fig_3}.

The critic network aims to reduce the time difference error or the prediction error, and its optimization target is

\begin{equation} \label{eq_4}
    L^V(\phi)=\hat{E}[\hat{A_t}]
\end{equation}
where $\phi$ is the critic-network parameter, $\hat{E}[\cdot]$ the estimated empirical mean of small batch data, and $\hat{A_t}$ the advantage function for estimating the error of time differences. Generalized Advantage Estimation~(GAE) is often used to calculate $\hat{A_t}$ at time step $t$, i.e., 
\begin{equation} \label{eq_5}
    \hat{A_t}=\delta_t+(\gamma\lambda)\delta_{t+1}^V+(\gamma\lambda)^2\delta_{t+2}^V\cdot\cdot\cdot+(\gamma\lambda)^{\mathcal{B}-t-1}\delta_{\mathcal{B}-1}^V
\end{equation}
where $\lambda\in[0,1]$ is the GAE parameter, $\mathcal{B}$ the sample size of the small batch data. $\delta_t$ can be defined as
\begin{equation} \label{eq_6}
    \delta_t=r_t+\gamma V_\phi(s_{t+1})-V_\phi(s_t).
\end{equation}

Finally, the parameters of the critic network $V_\phi$ can be updated using stochastic gradient descent as shown below
\begin{equation} \label{eq_7}
    \phi=\phi-\eta_\phi\nabla L^V(\phi)
\end{equation}
where $\eta_\phi$ denotes the learning rate of the critic-network optimizer.

The actor network here is for obtaining the gradient update step of the policy network. More specifically, the actor network is modified for every new policy, by using the old policy $\pi_{old}(a_t|s_t)$ and an advantage function $\hat{A_t}$ with an optimization target of
\begin{equation} \label{eq_8}
    L^{\text{clip}}(\theta)=\hat{E}[\min(r_t(\theta)),\text{clip}(r_t(\theta),1-\varepsilon,1+\varepsilon)\hat{A_t}]
\end{equation}
where $\theta$ is an actor-network parameter, $r_t(\theta)=\frac{\pi_\theta(a_t|s_t)}{\pi_{old}(a_t|s_t)}$, and $\varepsilon$ a small hyperparameter to set a constrain on the distance between the new policy and the old policy, making the actor policy network coverage better. 

Similarly, the parameters of the actor network $\pi_\theta$ can be updated using stochastic gradient descent
\begin{equation} \label{eq_9}
    \theta=\theta-\eta_\theta\nabla L^{\text{clip}}(\theta)
\end{equation}
where $\eta_\theta$ denotes the learning rate of the actor network optimizer.

\subsection{Human Driving Prior Modeling}

Up to here, we can use PPO algorithms to generate continuous controls of AVs, but the naturalness of driving behaviors cannot be fully guaranteed. To solve this and have a truly human driving prior, we further introduce GAIL~\cite{ho2016generative} to supervise the training process. It is tempting to use traditional imitation learning solutions~\cite{fang2019survey} directly by having behavioral cloning through training on expert experience data. However, we notice that the traditional solution may lead to overfitting problems, making the model hard to be generalized, and when reinforcement learning tasks that require temporal decision-making are involved, compound errors become unavoidable. Put it simply, behavioral cloning can mimic the training dataset well while it is vulnerable to variations of unseen situations. Furthermore, behavioral cloning generates each state-action pair independently without considering the relations between pairs, and therefore a small error in early steps can lead to unbearable errors down the line. We also notice that Inverse Reinforcement Learning~(IRL) comes after the imitation learning solution with better performance~\cite{arora2021survey}, but the computational cost is often high and the learning efficiency is relatively lower. GAIL, on the other hand, builds upon Generative Adversarial Network~(GAN)~\cite{goodfellow2020generative} and IRL, which learns policy directly from expert trajectories, and more importantly, it significantly improves learning efficiency and is easy to be generalized. Similar to GAN, GAIL consists of a generator $G_{\phi}$ and a discriminator $D_{\xi}$. The generator models the human expert driving behavior and outputs the action $a_t$ based on the input state $s_t$. The discriminator accepts both the generator state-action pair and the expert trajectory state-action pair, i.e., $(s_{\phi,t}, a_{\phi,t})$, $(s_{_{E,t}}, a_{_{E,t}})$ respectively, as the input, and outputs a real number from 0 to 1 to discriminate whether the input state-action pair is from the generator or the human expert. By doing this, GAIL can use these two networks to correctly generate pairs, making it robust to the overfitting problem, i.e., robust to unseen situations. Moreover, in GAIL, the discriminator compares the entire trajectory generated by the policy against the expert trajectory, instead of the individual state-action pair, making it immune from compound errors.

When we train GAIL, the goal of the discriminator $D_{\epsilon}$ is to maximize the classification accuracy between the generator policy and the expert policy, which can be expressed as

\begin{equation} \label{eq_10}
\begin{aligned}
    &\mathcal{L}(\varphi,\xi)=\underset{a_{\varphi,t}\sim G_{\varphi}(\cdot|s_{\varphi,t})}{E}[\log D_{\xi}(s_{\varphi,t},a_{\varphi,t})]+\\
    &\hspace{4.3em}\underset{a_{E,t}\sim \pi_{E}(\cdot|s_{E,t})}{E}[\log(1-D_{\xi}(s_{E,t},a_{E,t})]\\
\end{aligned}
\end{equation}
where $\phi$ is the parameter of generator network, and $\xi$ the parameter of discriminator network.

Meanwhile, the goal of the generator is to make the generated trajectories misclassified as expert trajectories by the discriminator, so the output of the discriminator $D_{\xi}$ can be used as a reward function to train the generator policy. With that in mind, we use the PPO algorithm as the generator of the GAIL algorithm and design the reward function as
\begin{equation} \label{eq_11}
    r(S_{\phi,t},a_{\phi,t};\xi)=-\log(D_{\xi}(S_{\phi,t},a_{\phi,t}))
\end{equation}

After alternately training the generator and the discriminator in an adversary manner, the data distribution between the generated one and the real expert one becomes indistinguishable, leading to the needed human Driving priors. We present the detailed network structure in the following subsection. It is important to understand that the previously introduced PPO algorithm is for generating trajectories from a predefined state-action space, and its critic network is for penalizing the reward function if the trajectory deviates from the space. On the other hand, the discriminator network in the context of GAIL is responsible for penalizing trajectories that deviate from the expert's trajectories, i.e., from the real-world dataset, thus preserving the naturalness of generated trajectories.

\subsection{PPO and GAIL Network Structure Design}

\begin{figure}
\centering
\includegraphics[width=3.5in]{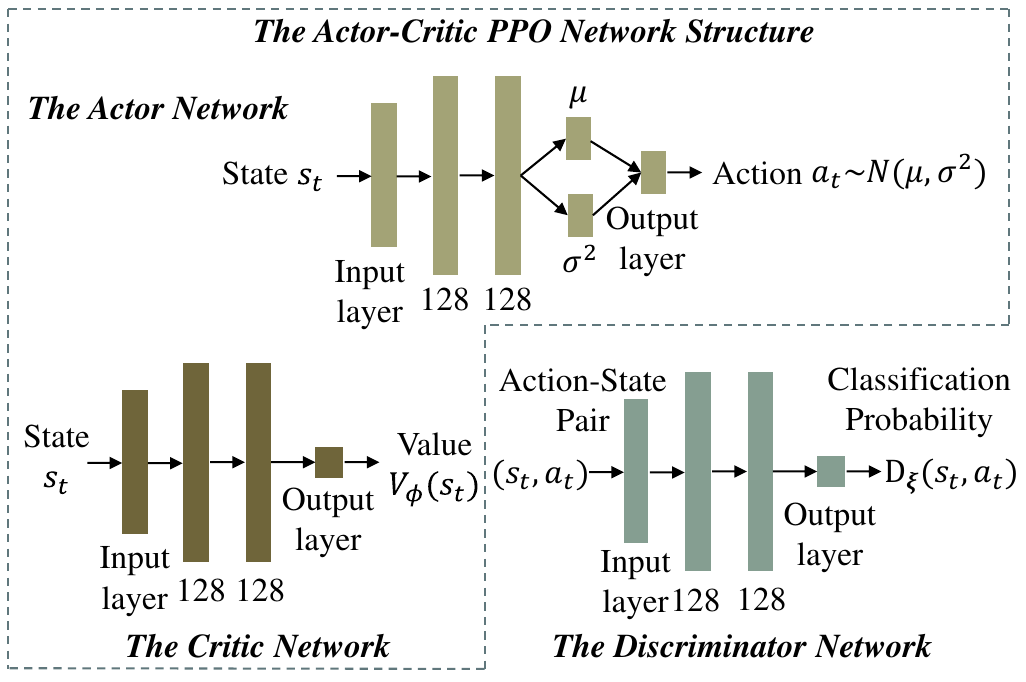}
\caption{The network structure of PPO-enabled GAIL.}
\label{fig_4}
\end{figure}

Recall that overfitting and training efficiency are the main concerns for traditional networks. In solving this, given the low dimensionality in the agent state space, we design a lightweight network structure by using one fully connected layer and a ReLU activation function. We present the network structure in Fig.~\ref{fig_4} in which PPO builds upon an actor network and a critic network, and GAIL consists of the PPO network as the generator and a discriminator.

For the actor network, the input state $s_t$, first passes through an input layer with a number of neurons equal to the dimension of $s_t$, and then passes through two hidden layers with 128 neurons, and the results will be further decomposed into two head networks to generate the mean $\mu$ and variance $\sigma^2$ of a Gaussian distribution, and the output layer samples the Gaussian distribution and outputs the action $a_t$. For the critic network, with the same input, the input layer, and the hidden layer as the actor network, its output layer outputs an estimated $V_{\phi}(s_t)$ to the current state $s_t$. Finally, sharing the same network with the critic network, the discriminator network uses the state-action pair as the input, i.e., the concatenation of $s_t$ and $a_t$, and the classification probability, $D_{\xi}(s_t,a_t)$, of the state-action pair, as the output.

\subsection{Natural Adversarial Policy Modeling and the Design of its Reward Functions}

We construct the natural adversarial policy model according to the PPO algorithm and the trained GAIL model as discussed in the previous subsections. To guarantee both the adversariality and naturalness, the key is to design the related reward functions that can be used to train the agent, i.e., adversariality reward and naturalness reward, and henceforth we explain our design as follows.

\subsubsection{Adversariality Reward}

We use the adversarial reward for a trained agent to generate disturbing actions to the AV under test, e.g., emergency braking, and sudden lane changes. The adversarial reward at time step $t$ can be expressed as
\begin{equation} \label{eq_12}
    \mathcal{R}_{\text{adv},t}=r_{d,t}(p_{\text{AV},t_0},p_{\text{agent},t_0},p_{\text{AV},t},p_{\text{agent},t})+r_{c,t}
\end{equation}
where $p_{\text{AV},t_0}$ and $p_{\text{agent},t_0}$, and $p_{\text{AV},t}$ and $p_{\text{agent},t}$ denote the position of the AV under test and the trained agent at the moment of initialization $t_0$, and at time step $t$, respectively. $r_{d,t}$ as a metric refers to the distance between the AV under test and the trained agent. A smaller $r_{d,t}$ means a collision is likely to happen, namely more danger, and then more rewards, which can be formulated as
\begin{equation} \label{eq_13}
\begin{aligned}
    & r_{d,t}(p_{\text{AV},t_0},p_{\text{agent},t_0},p_{\text{AV},t},p_{\text{agent},t})=\\
    & \text{clip}(\frac{\|p_{\text{AV},t_0}-p_{\text{agent},t_0}\|_2-\|p_{\text{AV},t}-p_{\text{agent},t}\|_2}{\|p_{\text{AV},t_0}-p_{\text{agent},t_0}\|_2},-1,1)\\
\end{aligned}
\end{equation}
The clip function used here is a convenient tool in learning networks to set a threshold for each updated policy, and by doing this, we can stabilize the learning procedure, and avoid large and harmful policy updates.

Note that in the simulation environment, since the AV under test and the agent cannot penetrate each other, their L2 distance will always be greater than 0, i.e., $r_{d,t} < 1$, making a weak feedback signal to collisions. To highlight the collision reward, we introduce $r_{c,t}$ in Eq.~\ref{eq_13} as the collision reward. By doing this, if an agent collides with the AV under test, it will be rewarded, otherwise, if an agent collides with other vehicles, it will be penalized. In detail, $r_{c,t}$ can be represented as

\begin{equation} \label{eq_14}
\
    r_{c,t}=\left\{
                \begin{array}{ll}
                   \hspace{0.8em}1, \hspace{0.8em} \text{collided\ with\ the\ AV\ under\ test},\\
                   \hspace{0.8em}0,  \hspace{0.8em} \text{no\ collision},\\
                  -1, \hspace{0.8em} \text{collided\ with\ other\ vehicles.}
                \end{array}
              \right.
\
\end{equation}

\subsubsection{Naturalness Reward}
To guarantee naturalness, the pre-trained human driving prior model GAIL is in use to supervise the naturalness of the agent's behavior. We notice that both the action output of the agent and the GAIL prior model follow multivariate Gaussian distributions, and therefore, we decide to use Kullback-Leibler~(KL) divergence~\cite{malinin2019reverse} to measure the distance between these two distributions. The naturalness reward can be expressed as
\begin{equation} \label{eq_15}
    \mathcal{R}_{nat,t}(s_t)=\text{clip}(\frac{M-\text{KL}(G_{\phi}(\cdot|s_t),\pi_{\theta}(\cdot|s_t))}{M},0,1)
\end{equation}
where $G_{\phi}(\cdot|s_t)$ and $\pi_{\theta}(\cdot|s_t)$ denote the action distribution output of the GAIL generator and the agent at state $s_t$, respectively. For KL divergence, after assuming $G_{\phi}(\cdot|s_t)$ is the real human driving behavior distribution, it needs to make sure that the fitted distribution $\pi_{\theta}(\cdot|s_t)$ is getting closer to $G_{\phi}(\cdot|s_t)$. $M$ is an empirical value for the KL divergence of an agent at the beginning of training. Finally, based on Eq.~\ref{eq_12} and Eq.~\ref{eq_15}, we can conclude the natural adversarial reward function as
\begin{equation} \label{eq_16}
    \mathcal{R}_t=\mathcal{R}_{adv,t}+\mathtt{\varpi}\mathcal{R}_{nat,t}
\end{equation}
where $\mathtt{\varpi}$ is the weight factor for balancing between the adversariality and naturalness.

\section{Experiment}
\label{experiment}

\subsection{Experimental setup}
\subsubsection{Human Driving Traffic Dataset}

\begin{figure}
  \centering
  \begin{subfigure}{0.43\linewidth}
    \includegraphics[width=\linewidth]{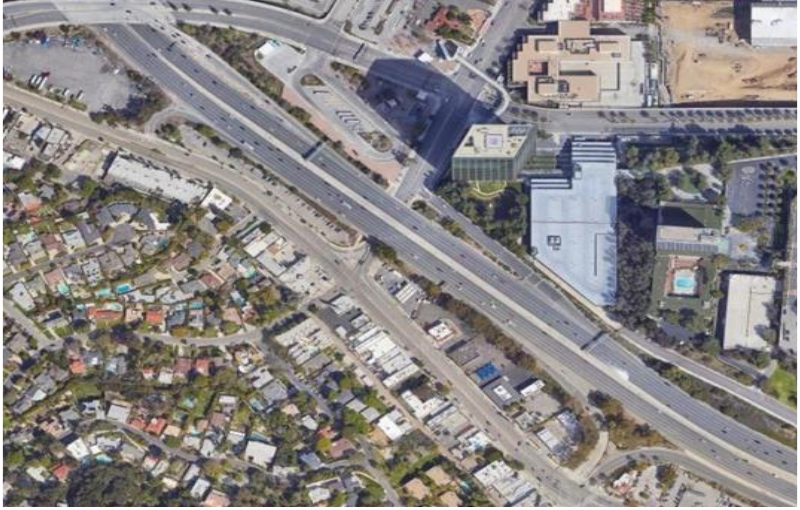}
    \caption{NGSIM@Google map.}
    \label{fig:sub1}
  \end{subfigure}
  \begin{subfigure}{0.45\linewidth}
    \includegraphics[width=\linewidth]{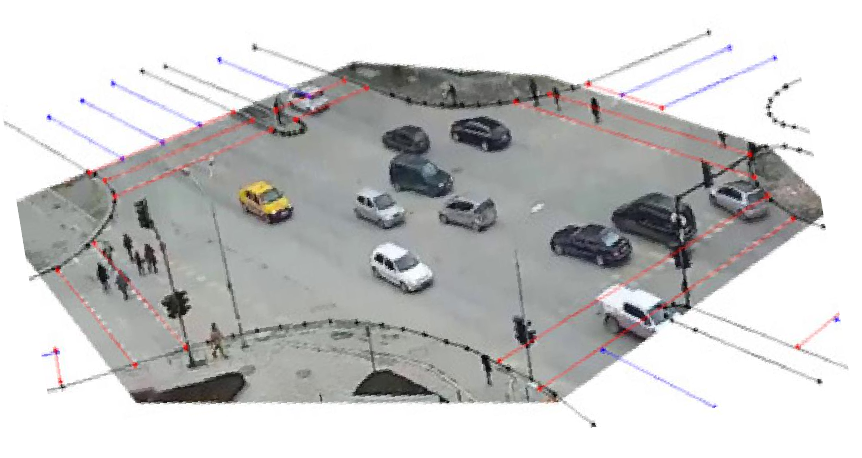}
    \caption{INTERACTION@\cite{zhan2019interaction}}.
    \label{fig:sub2}
  \end{subfigure}
  \begin{subfigure}{0.455\linewidth}
    \includegraphics[width=\linewidth]{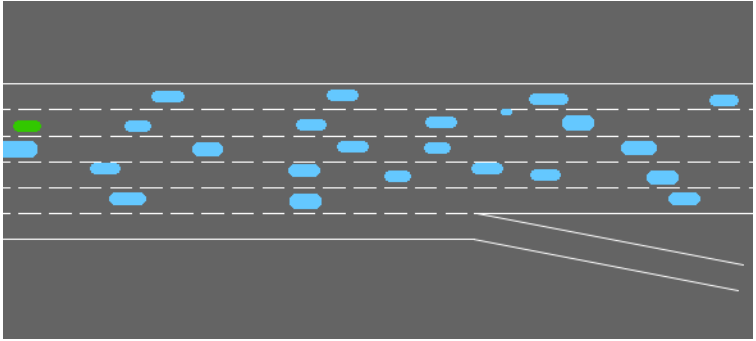}
    \caption{Road of NGSIM.}
    \label{fig:sub3}
  \end{subfigure}
  \begin{subfigure}{0.45\linewidth}
    \includegraphics[width=\linewidth]{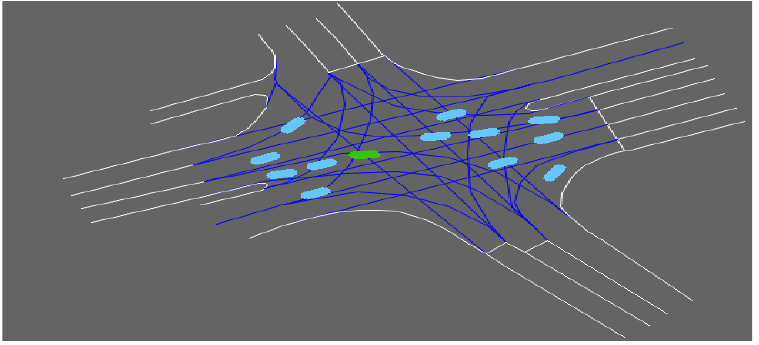}
    \caption{Road of INTERACTION.}
    \label{fig:sub4}
  \end{subfigure}
  \caption{Constructing road structures from the real world datasets.}
  \label{fig:road}
\end{figure}

The experiments have been conducted on public datasets NGSIM~\cite{leurent2018environment} and INTERACTION~\cite{zhan2019interaction}. The two datasets were selected primarily because of their complex interactions among vehicles in both highway and urban settings. {\bf NGSIM:} This dataset includes trajectories of vehicles on US-101, I-80, etc., while the data was collected every one-tenth of a second to precisely monitor the vehicles' movements. We use the dataset from a 640m long highway located in US-101, including five mainline lanes, and a sixth auxiliary lane for on-ramp and off-ramp, separately.
{\bf INTERACTION:} This dataset includes trajectories from various junctions across several countries. We utilize the subset labeled TC\_BGR\_Intersection\_VA, where vehicles frequently interact with other vehicles and pedestrians.

\subsubsection{Traffic Simulation Environment Construction}

\begin{figure}
  \centering
  \begin{subfigure}{\linewidth}
    \includegraphics[width=\linewidth]{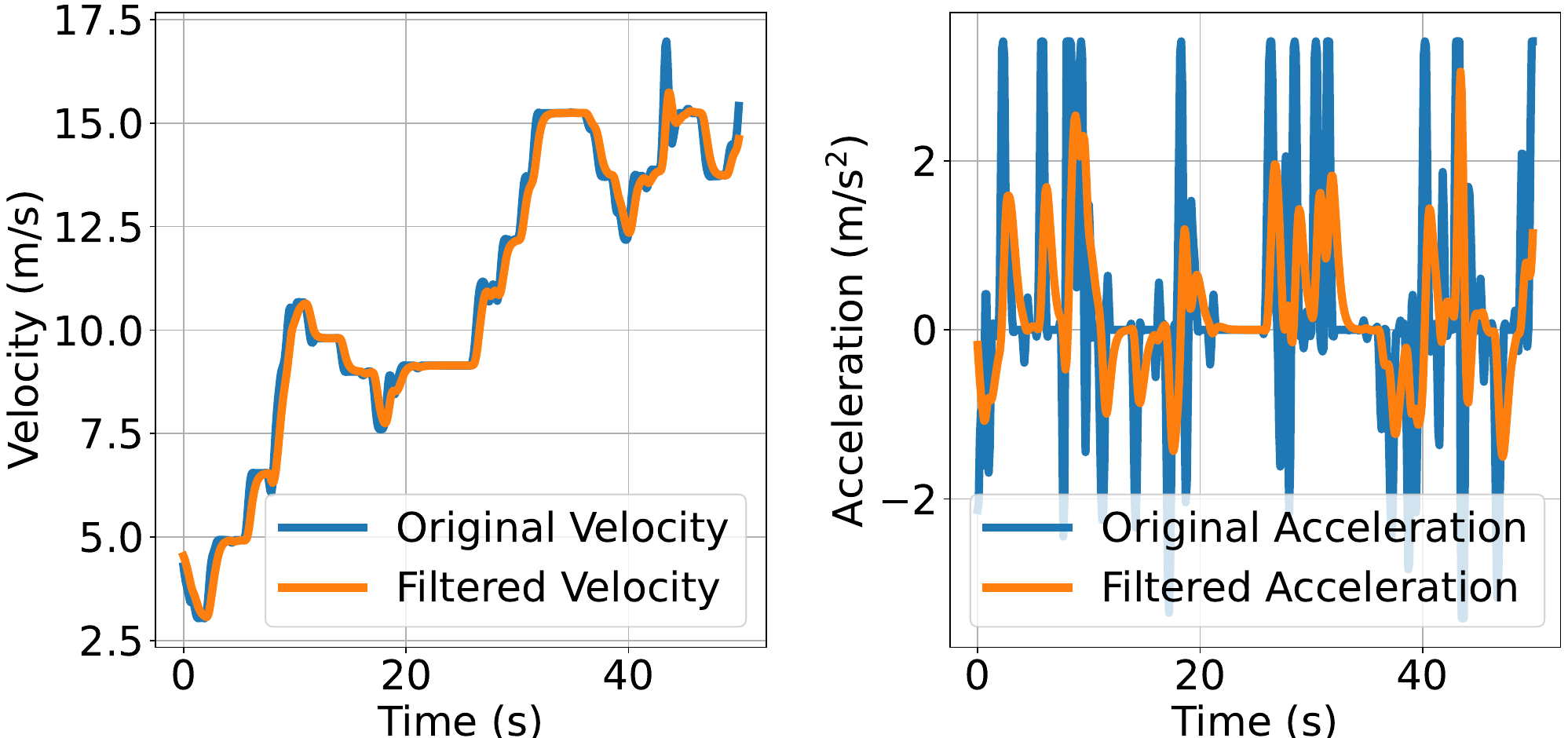}
    \caption{IDM.}
    \label{fig:filter_IDM}
  \end{subfigure}
  \begin{subfigure}{\linewidth}
    \includegraphics[width=\linewidth]{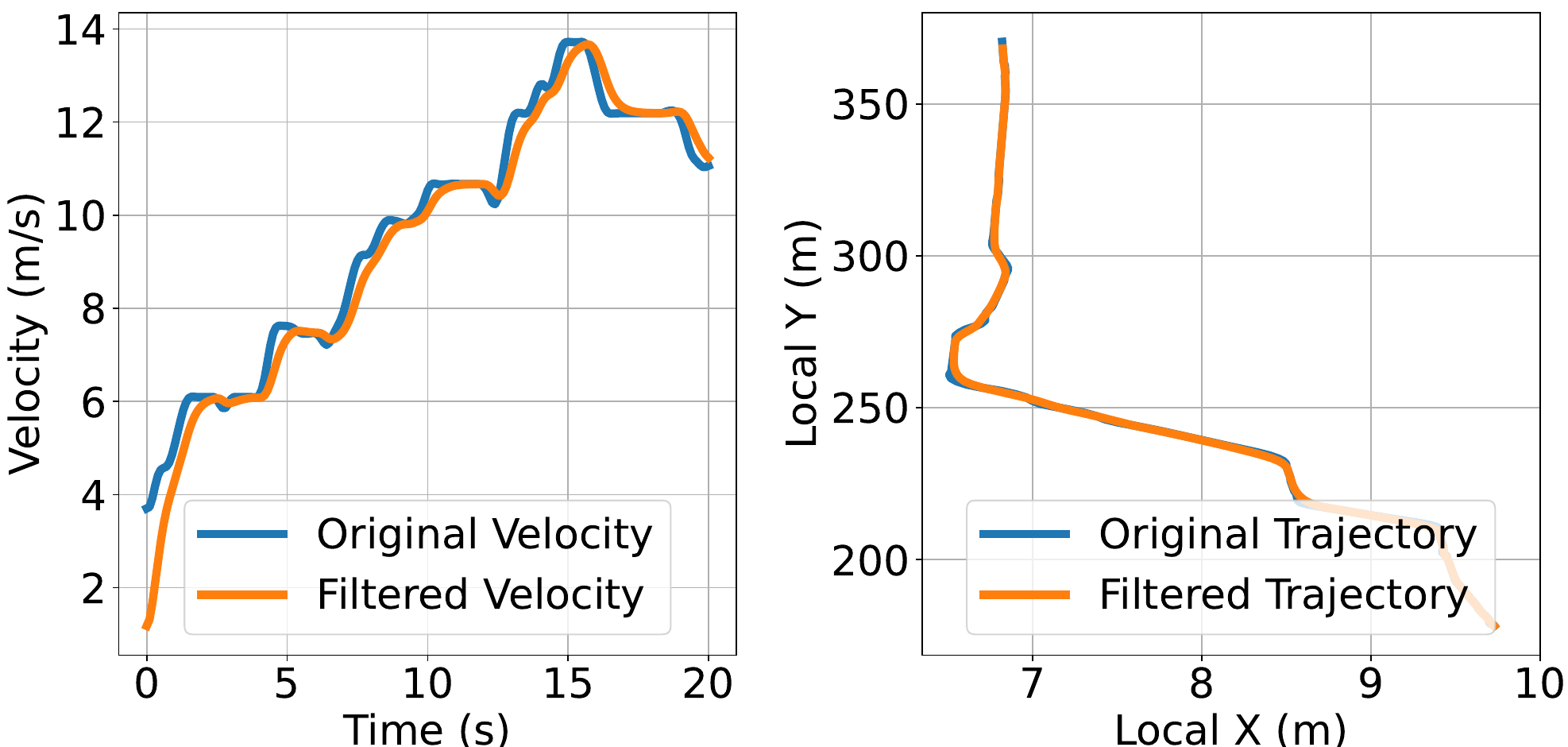}
    \caption{MOBIL.}
    \label{fig:filter_MOBILE}
  \end{subfigure}
  \caption{Applying the sEMA filter for data preprocessing.}
  \label{fig:filter}
\end{figure}

Similar to existing work for driving behavior modeling~\cite{kuefler2017imitating,huang2021driving}, we build the traffic simulation environment on Highway-env, a lightweight platform designed for generating traffic scenarios. Before utilizing the above-mentioned datasets, we first need to adapt the road structure of the dataset to be compatible with the Highway-env simulator. While the highway road in NGSIM can be directly transferred to the simulator with minimum alterations, junctions in INTERACTION often feature more irregular road structures, requiring further adjustments. These adjustments include fine-tuning the coordinates and calculating the length of arc curves. Once these adjustments are complete, as shown in Fig.~\ref{fig:road}, the adapted road structure can be introduced to the simulator. In addition, as stated in Algo.~\ref{alg:IDM}, preprocessing of the datasets is required to select qualified scenarios and remove the noise. Specifically, for NGSIM, we initially identified 20,443 car-following scenarios, from which 5,034 were selected. For INTERACTION, 8,603 car-following scenarios were identified, and 127 were selected. For the chosen scenarios, we employed the sEMA filter to remove the noise of velocity and acceleration for IDM, and velocity and trajectories for MOBIL. The exemplary results of this noise reduction are illustrated in Fig.~\ref{fig:filter}.

\begin{figure}
    \centering
    \includegraphics[width=.8\linewidth]{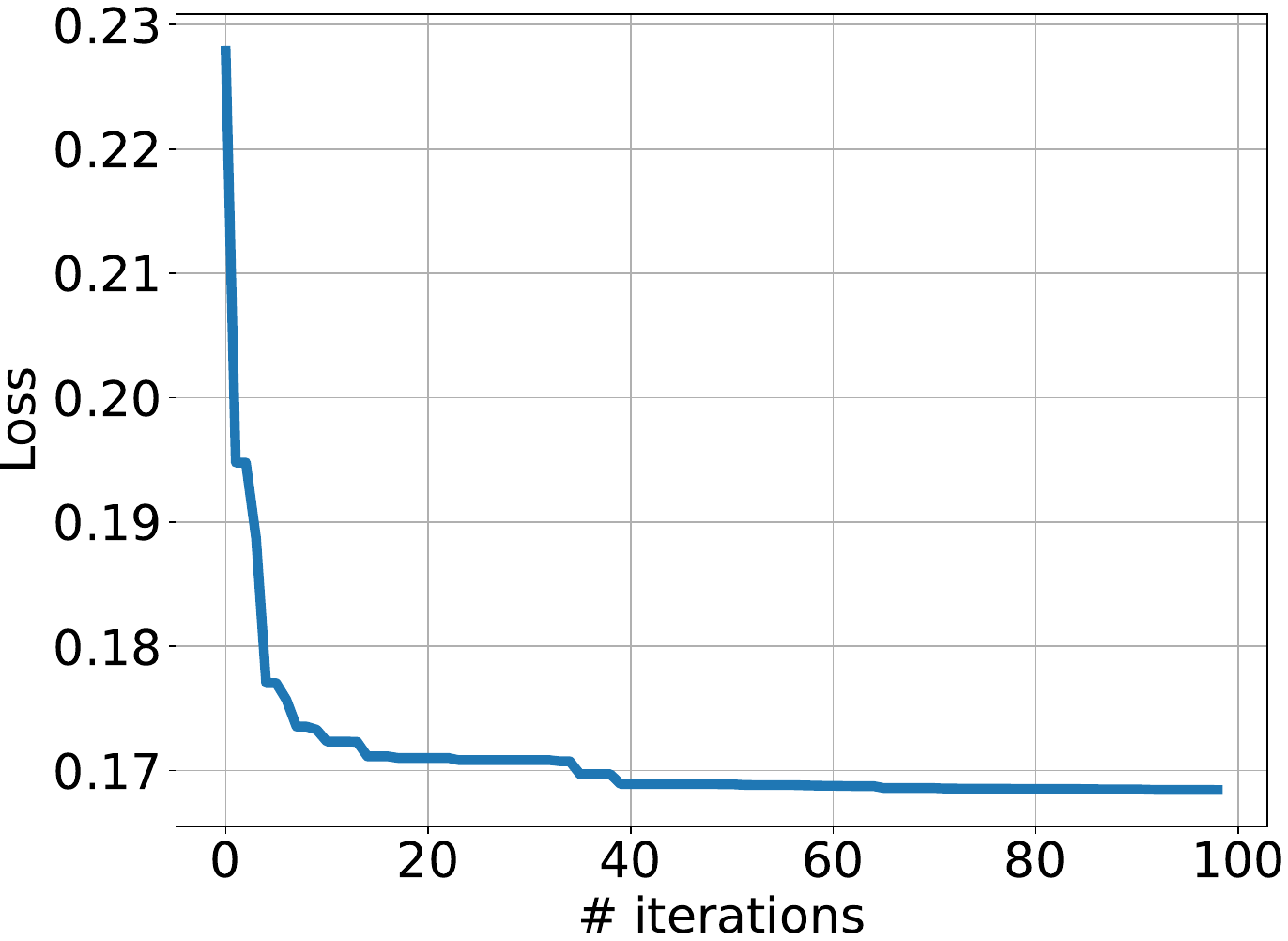}
    \caption{Iteration loss of the genetic optimizer.}
    \label{fig:genetic}
\end{figure}

\begin{figure}
    \centering
    \includegraphics[width=\linewidth]{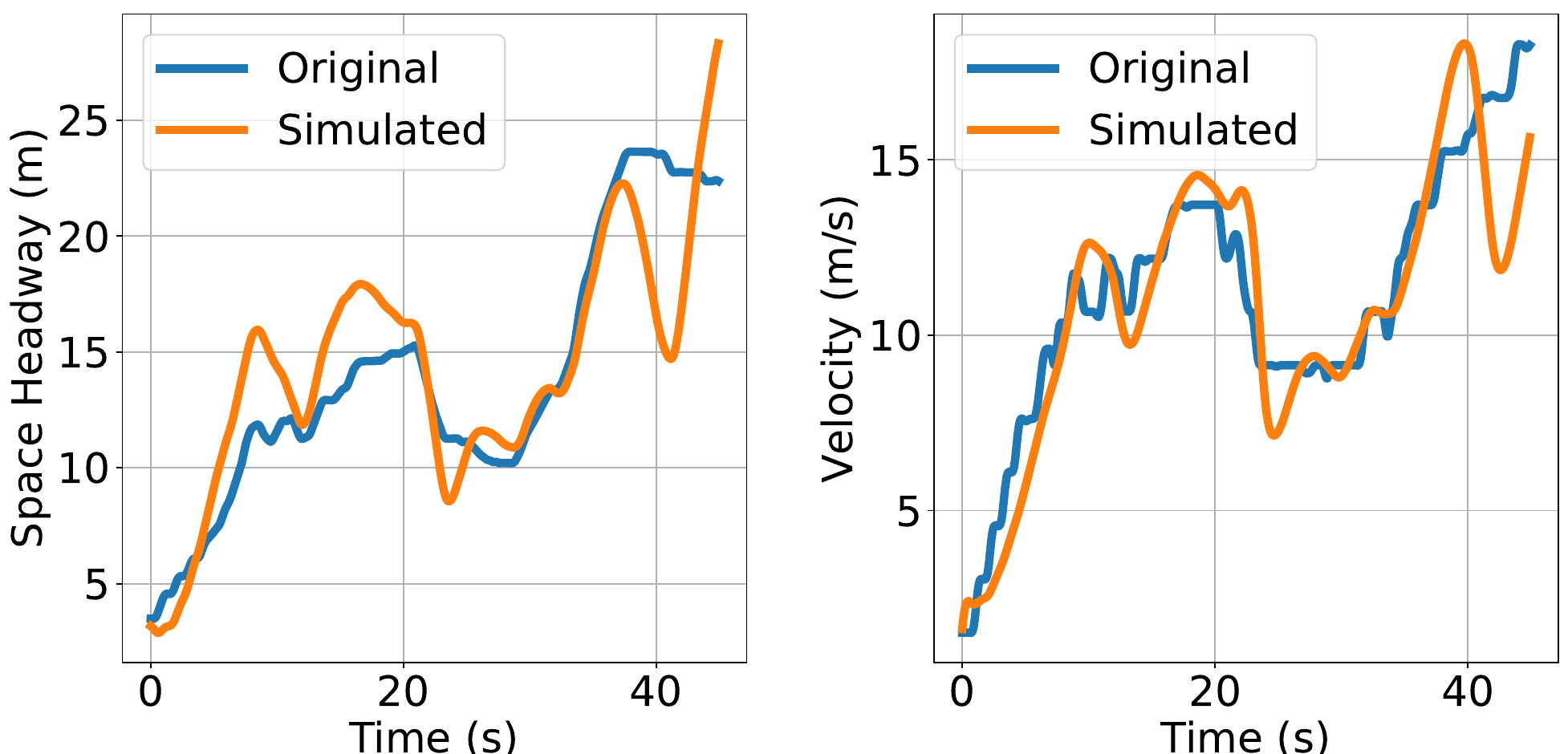}
    \caption{The comparison between IDM simulated results and the original data.}
    \label{fig:IDM_compare}
\end{figure}

Next, we can calibrate the parameters of IDM and MOBIL using the method described in Sec.~\ref{design}. For IDM, we start by setting ranges of its parameters using empirical knowledge~\cite{kesting2008calibrating}. The parameters include: $a_{_\text{IDM}}$ representing the maximum acceleration; $v_{_\text{IDM}}$, the desired speed; $\delta_{_\text{IDM}}$, the acceleration exponent; $s_{_\text{IDM}}$, the desired following distance; $b_{_\text{IDM}}$, the comfortable deceleration; and $T$, the safe headway time distance. Based on these ranges, we employ the genetic optimizer, as illustrated in Sec.~\ref{design}, to calibrate the parameters. The range and the calibrated parameter are presented in Table.~\ref{tab:IDM_parameter}. Note that $\delta_{_\text{IDM}}$ represents the aggressiveness of a driver, which often has 4 as its default value. As shown in Fig.~\ref{fig:genetic}, the genetic optimizer is well converged to an acceptable loss at 0.168 within 100 rounds of iterations. A comparison between the original measurements of space headway and velocity, and the result simulated using IDM, is depicted in Fig.~\ref{fig:IDM_compare}. This comparison illustrates a close match between the IDM simulated data and the original observations, thereby validating our parameter calibration schemes.

\begin{figure}
    \centering
    \includegraphics[width=.8\linewidth]{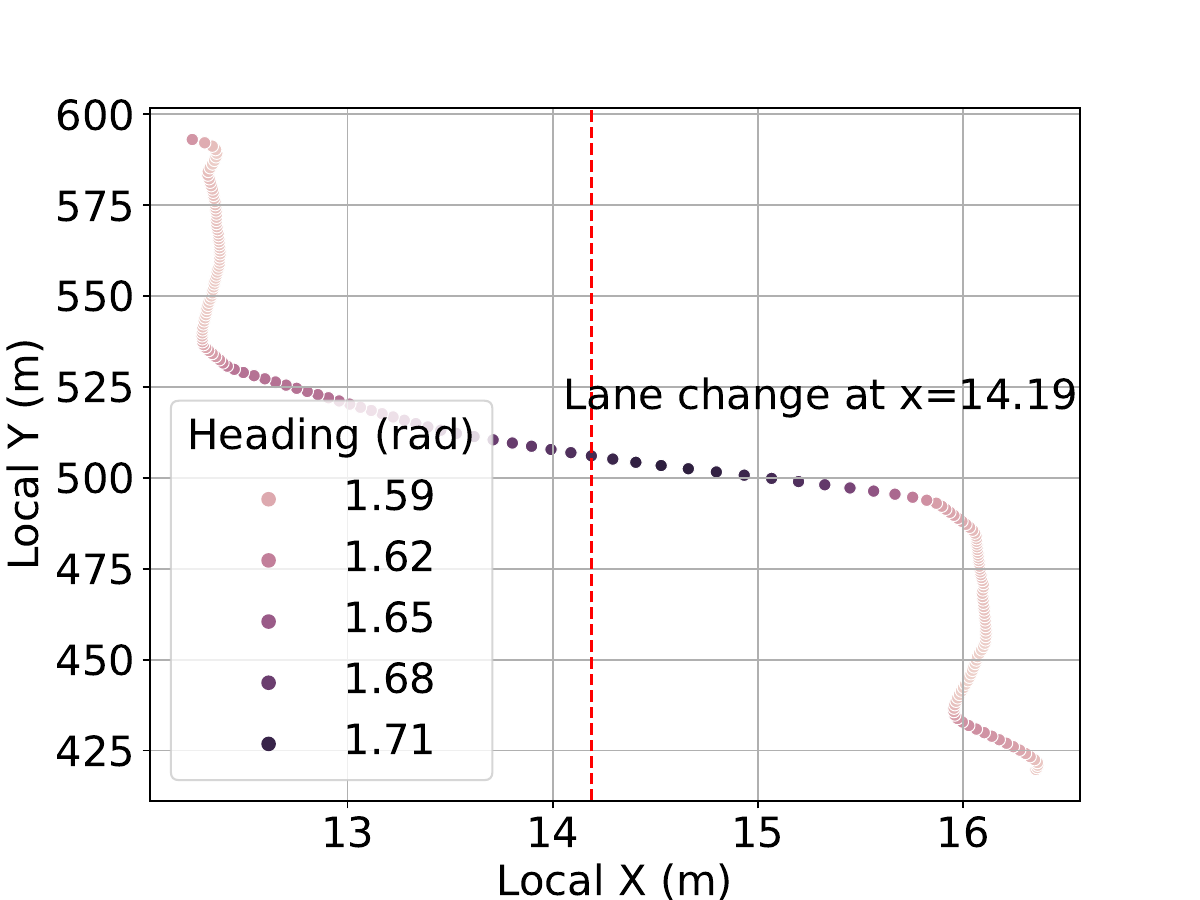}
    \caption{Capturing the trajectory that has the greatest heading.}
    \label{fig:MOBIL_heading}
\end{figure}

For MOBIL, the parameters are typically calibrated by statistical analysis from real-world data and existing studies. In detail, to enhance the accuracy, we need to calibrate MOBIL-related parameters precisely when the vehicle is changing lanes. To achieve this, we capture the trajectory that has the greatest heading, as illustrated in Fig.~\ref{fig:MOBIL_heading}. After that, we can obtain the calibrated parameters from the captured trajectories: $\Delta a_{\mathrm{th}}=0.2\unit{m/s^2}$, representing the acceleration gain, and $\text{MAX\_BRAKING\_IMPOSED}=2\unit{m/s^2}$, representing the maximum braking imposed to the follower, in line with the existing work~\cite{kesting2007general,wang2022evaluation,huang2021driving}.

\begin{table}
\caption{Parameter calibration of the IDM Model.}
\centering
\begin{tabular}{c|c|c|l}
\toprule
{\bf Parameter} & {\bf Range} & {\bf \makecell{Calibrated\\Result} } & {\bf Unit}\\
\midrule
$a_{_\text{IDM}}$ & $[0.1, 6]$ & 2 & $\unit{m/s^2}$\\
$v_{_\text{IDM}}$ & $[1, 70]$ & 10 & $\unit{m/s}$\\
$\delta_{_\text{IDM}}$ & --- & 4 & --- \\
$s_{_\text{IDM}}$ & $[0.1,8]$ & 1 & $\unit{m}$\\
$b_{_\text{IDM}}$ & $[0.1,6]$ & 1 & $\unit{m/s^2}$ \\
$T$ & $[0.1, 5]$ & 0.5 & $\unit{s}$ \\
\bottomrule
\end{tabular}
\label{tab:IDM_parameter}
\end{table}

\begin{figure}
  \centering
  \begin{subfigure}{.44\linewidth}
    \includegraphics[width=\linewidth]{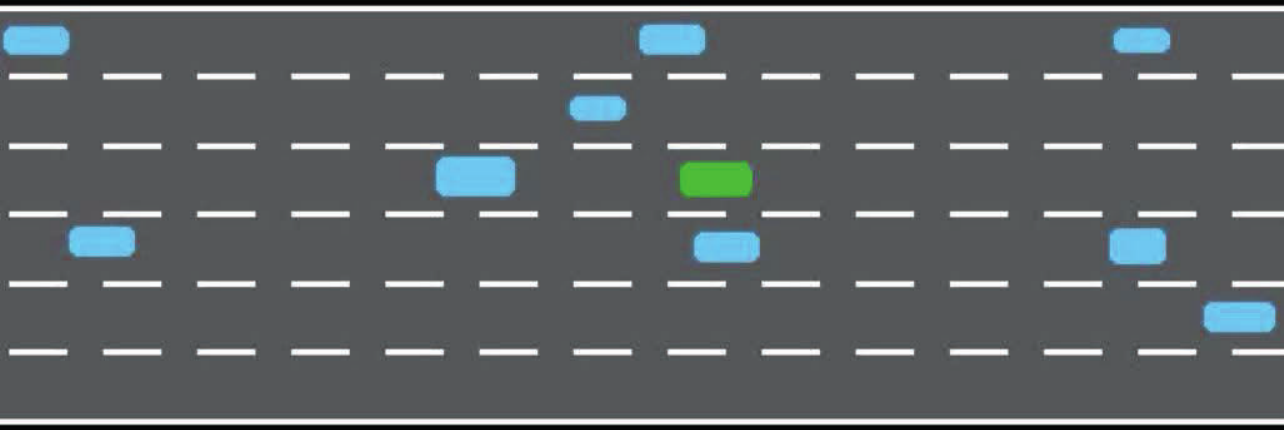}
    \caption{The training env. for GAIL.}
    \label{fig:GAIL_train}
  \end{subfigure}
  \begin{subfigure}{.45\linewidth}
    \includegraphics[width=\linewidth]{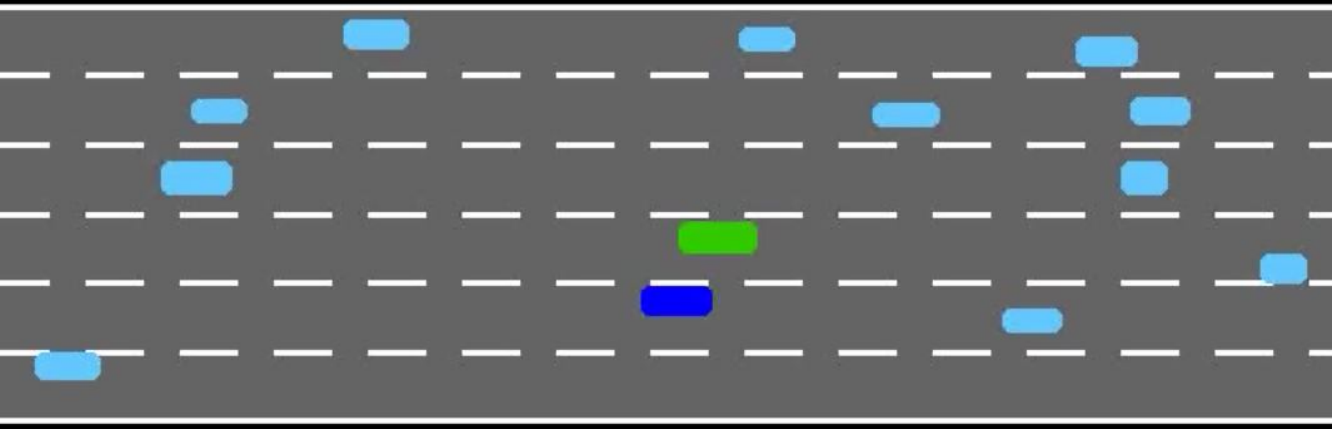}
    \caption{The training env. for PPO.}
    \label{fig:PPO_train}
  \end{subfigure}
  \caption{Traffic simulation using Highway-env simulator.}
    \label{fig:highway_train}
\end{figure}

In the simulation environment settings, we use different schemes for GAIL and PPO. Specifically, for GAIL, we initialize the environment with a randomly selected ego vehicle, and the action of this ego vehicle is controlled by the GAIL policy network. Other vehicles in the background do not respond to any behaviors of the ego vehicle and drive according to their predefined trajectories. By defining it like this, we can make sure the GAIL policy follows human driving priors better. A snapshot of traffic simulation for GAIL is shown in Fig.~\ref{fig:GAIL_train}, where the green vehicle is the ego vehicle, and the light blue vehicles are the background vehicles. For PPO, the ego vehicle is randomly selected as well and is controlled by the PPO policy network. The difference is that for the selection of the AV under test, a search for all neighbor vehicles within a range of 50 meters~\footnote{According to the dataset we use, there must be a vehicle within the range of 50 meters~\cite{huang2021driving}.} will be conducted first, and the closest vehicle in the same lane or adjacent lane to the ego vehicle will be selected as the AV under test. If that search fails, the closest vehicle to the ego vehicle but outside the target lane will be selected as the AV under test. All vehicles except for the ego vehicle are controlled in a surrogate model, which means they can respond to their surrounding vehicles, e.g., active brakes, lane changes, etc. Meanwhile, an example of PPO simulation is shown in Fig.~\ref{fig:PPO_train}, where the green vehicle is the ego vehicle, the dark blue vehicle the AV under test, and the light blue vehicle the background vehicle.

\subsubsection{Model Training Details}
The supervision of GAIL's training demands human expert trajectories. We acquire these trajectories through the simulation environment with a sampling frequency of 10\unit{Hz}, and they appear as state-action pairs. The state is a feature vector of 56 dimensions, including 6 features of the length and width of the vehicle, the lateral offset of the lane centerline, the lateral speed, the longitudinal speed, and the steering, and the other 50 features for representing the relative information to the 10 nearest vehicles within 50\unit{m}, e.g., the relative lateral distance, relative longitudinal distance, relative lateral speed, relative longitudinal speed, and relative steering between the vehicles. Note that if the number of neighboring vehicles is less than 10, the emptied features' values are denoted as 0. The action is a vector of 2 dimensions, including the acceleration and steering, where the acceleration interval is $[-5\unit{m/s^2}, 3\unit{m/s^2}]$ and the steering interval is $[-\pi/3\unit{rad}, \pi/3\unit{rad}]$. The collecting process is as follows.

\begin{enumerate}
    \item From the NGSIM dataset, we select vehicles that are changing lanes, and from the INTERSECTION dataset, we make random selections.
    \item For each selected vehicle, we use 100 sampled intervals as one scenario and 4 scenarios in total. If the running time of a selected vehicle is greater than the time of 4 scenarios, i.e., 40\unit{s}, we randomly pick 4 scenarios from the whole running time.
    \item To focus on general traffic situations, empirically, the simulation will be skipped if a scenario contains more than 40 vehicles or the average speed of vehicles in that scenario is less than 5\unit{m/s} at its initial state.
    \item The state-action pair would be appended to the list of an expert trajectory, except for those simulations stopped prematurely, e.g., a collision happens to the ego vehicle, the ego vehicle off the road, or the simulation lasts more than 10\unit{s}.
\end{enumerate}

The supervision of PPO's training, on the contrary, demands no expert trajectories, while only the input state and the output action need to be specified. For the input state, it is a feature vector with 10 dimensions, including the relative information for the ego vehicles and the AV under test, i.e., the relative lateral distance, relative longitudinal distance, relative lateral speed, relative longitudinal speed, and relative steering. For the output action, it is a feature vector with 2 dimensions, including acceleration and steering. To generate adversarial behaviors, we set no constraints on the action space.

The PyTorch~\cite{paszke2019pytorch} is in use to train the GAIL and PPO, and the Adam optimizer~\cite{kingma2014adam} is adopted to optimize the network parameter. After carefully fine-tuning the training parameters, we obtain the hyperparameter configurations, as shown in Table~\ref{tab_2}, aiming to achieve the best model performance.

\begin{table}
\caption{The training parameters of GAIL and PPO.}
\label{tab_2}
\centering
\begin{tabular}{c|c|c}
\toprule
\textbf{Parameter} & \textbf{Description} & \textbf{Value} \\ 
\midrule
$\gamma$ & Reward discount factor & 0.99 \\
$\lambda$ & GAE coefficient & 0.95 \\
$\epsilon$ & Policy constraint factor & 0.2 \\
$\eta_{_{\phi}}$ & Critic-network learning rate & 0.001 \\
$\eta_{_{\theta}}$ & Actor-network learning rate & 0.0001 \\
$\eta_{_{\xi}}$ & Discriminator network learning rate & 0.0001 \\
$\mathcal{B}$ & Sampling batch size & 2048 \\
$M$ & Initial KL divergence & 25 \\
$\mathtt{\varpi}$ & Natural adversarial balance factor & 0.02 \\
$E$ & Maximum training episodes & 500 \\
\bottomrule
\end{tabular}
\end{table}

\subsection{Result Analysis of the Natural Adversarial Scenario Generation}
\subsubsection{Result Analysis of the Model Training}

\begin{figure}
\centering
\includegraphics[width=.9\linewidth]{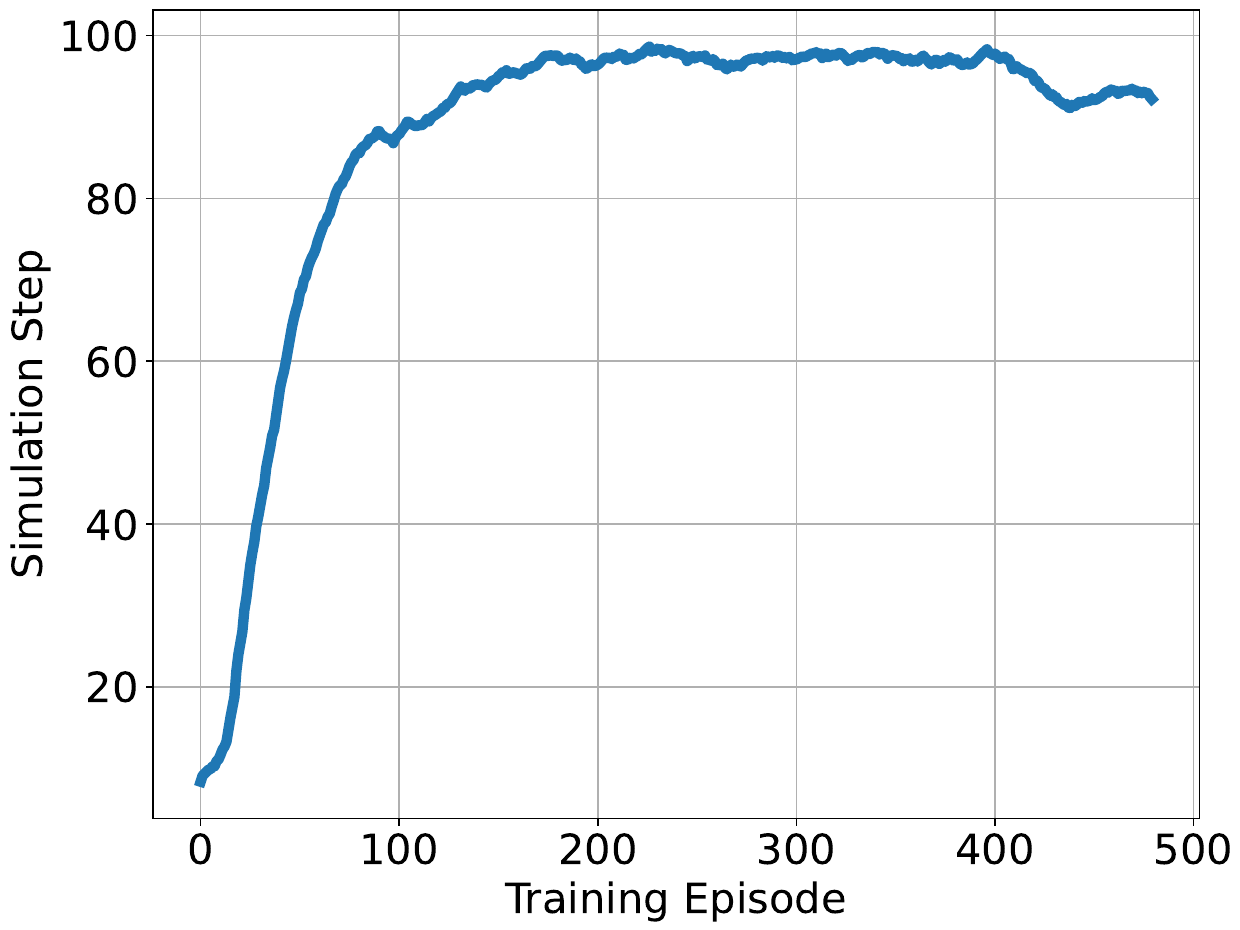}
\caption{The average simulation steps for different episodes.}
\label{fig_6}
\end{figure}

We begin by plotting the training episodes against the average simulation steps of the trained GAIL model, as shown in Fig.~\ref{fig_6}. After the 200th episode, the simulation steps stabilize, representing an acceptable performance in modeling the human driving prior. Specifically, the ego vehicle demonstrates smooth and natural driving with fewer unexpected collisions. Therefore we select the 200th episode for the supervised training of PPO. To evaluate the effectiveness of naturalness supervision, we compare two solutions: a baseline PPO model that considers only the adversarial reward function, as highlighted in pioneer work~\cite{chen2021adversarial,sun2021corner}, and our proposed method of natural adversarial training. Recent studies have overly emphasized the importance of adversarial actions, such as collisions, and consequently, naturalness reward was not incorporated into their function. Although traffic rules were considered to enhance the naturalness in some work~\cite{chen2021adversarial}, this approach only represents a border guideline. In contrast, our solution draws directly from real-world driving data, learning a policy that includes both adversariality reward and naturalness reward as discussed in Sec.~\ref{design}.

\begin{figure}
    \centering
    \includegraphics[width=\linewidth]{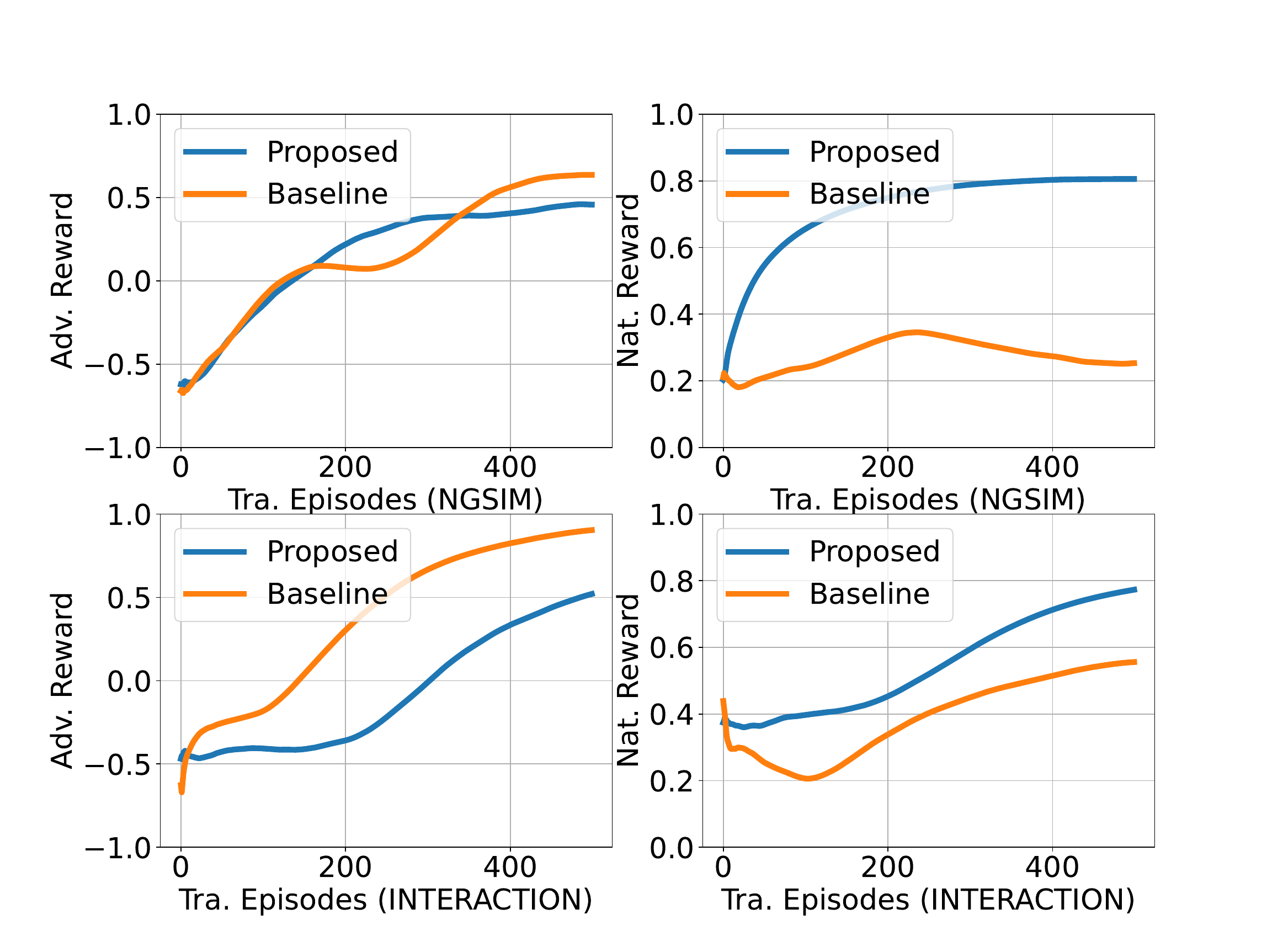}
    \caption{The adversariality and naturalness rewards comparison between the baseline and the proposed solutions.}
    \label{fig:reward_compare}
\end{figure}

Fig.~\ref{fig:reward_compare} displays the compared results, with the x-axis showing the number of training episodes and the y-axis showing the mean cumulative rewards across the two datasets, NGSIM and INTERACTION. For naturalness, our method achieved rewards of 0.81 and 0.77, surpassing the baseline's 0.25 and 0.56. Adversariality rewards were 0.46 and 0.52 after normalization, against the baseline's 0.64 and 0.9. 
\begin{equation}
    \mathbb{E} = w_N * N + w_A * A
    \label{eq:effectiveness}
\end{equation}
We employ a weighted sum model, as detailed in Eq.~\ref{eq:effectiveness}, to attain this balance, assigning different weights, $w_N$ and $w_A$, to naturalness~($N$) and adversariality~($A$) rewards. Generally, $w_N$ will be greater than $w_A$ to ensure that realistic scenarios guide the development of autonomous driving systems. However, in the worst case, naturalness and adversariality might share equal importance. With these considerations, our method's effectiveness is $\mathbb{E}_{\text{proposed}}^{\text{NGSIM}}=0.77$ and $\mathbb{E}_{\text{proposed}}^{\text{INTERACTION}}=0.765$, outperforming the baseline's $\mathbb{E}_{\text{baseline}}^{\text{NGSIM}}=0.535$ and $\mathbb{E}_{\text{baseline}}^{\text{INTERACTION}}=0.755$. These findings correspond to gains of 44\% for NGSIM and 1\% for INTERACTION. The takeaway message from our study contains two-fold: 1) our solution consistently outperforms the baseline in overall effectiveness, especially in naturalness, while still maintaining competitive levels of adversariality; 2) the improvement of our solution in INTERACTION is relatively minor. This can be attributed to the constrained action space in intersections, where movements are generally slower and lanes are fewer compared to highways. On the highway, vehicles can behave more aggressively, thus demanding a greater focus on naturalness.

\begin{figure}
    \centering
    \includegraphics[width=\linewidth]{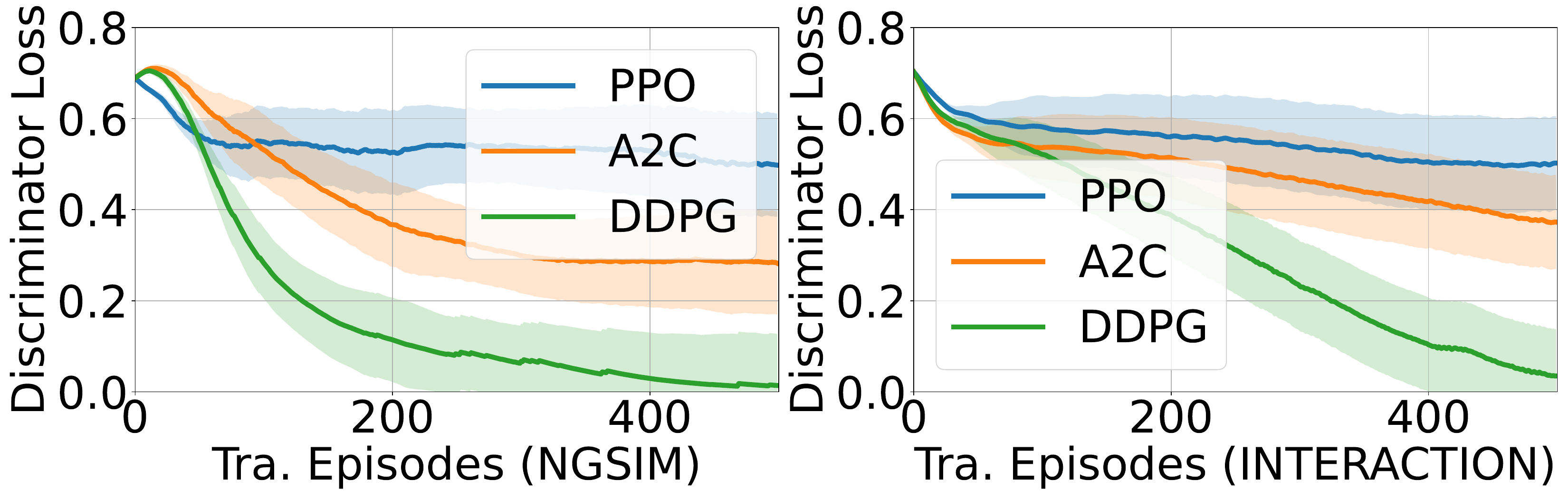}
    \caption{The discriminator loss comparison among PPO, A2C and DDPG.}
    \label{fig:PPO_compare}
\end{figure}
\subsubsection{Alternative Reinforcement Learning Algorithms}
Alternative reinforcement learning algorithms to PPO, such as A2C~(Advantage Actor-Critic)~\cite{mnih2016asynchronous} and DDPG~(Deep Deterministic Policy Gradients)~\cite{ddpg2016}, have also attracted much attention in the field. However, PPO proves to be superior in our context, owning to its clipping mechanism that enhances learning stability by preventing drastic policy updates. To understand the benefit of PPO's stability, we conducted experiments using the same setting as previously described and plotted the discriminator loss of these three algorithms in Fig.~\ref{fig:PPO_compare}. The plot reveals that PPO converges to a loss at 0.5, meaning that the discriminator cannot differentiate the original and generated data, thus achieving a Nash equilibrium. In contrast, the continuous decrease in the loss for DDPG and A2C indicates their inability to stabilize the policy updates, resulting in suboptimal outcomes. It's important to note that while the clipping mechanism promotes stability, it may also lead to suboptimal solutions. A trade-off is needed to balance the stability and the optimality, and in our study, we set the clipping parameter, or the policy constraint factor $\epsilon$, to 0.2, which provides the best results given our task.

\subsubsection{Statistical Analysis of the Key Metrics}

Next, we run 2000 rounds of simulations to generate test scenarios through the baseline model and the proposed natural adversarial model. A statistical analysis of the simulation results has been conducted via some key metrics, i.e., the collision rate and the number of lane changes for the macroscopic traffic metrics, and acceleration and steering for the microscopic traffic metrics. Note that for simplicity, the numerical analysis below is majorly from NGSIM, INTERACTION shares similar results as observed in our study.

\begin{figure}
  \centering
  \begin{subfigure}{\linewidth}
    \includegraphics[width=\linewidth]{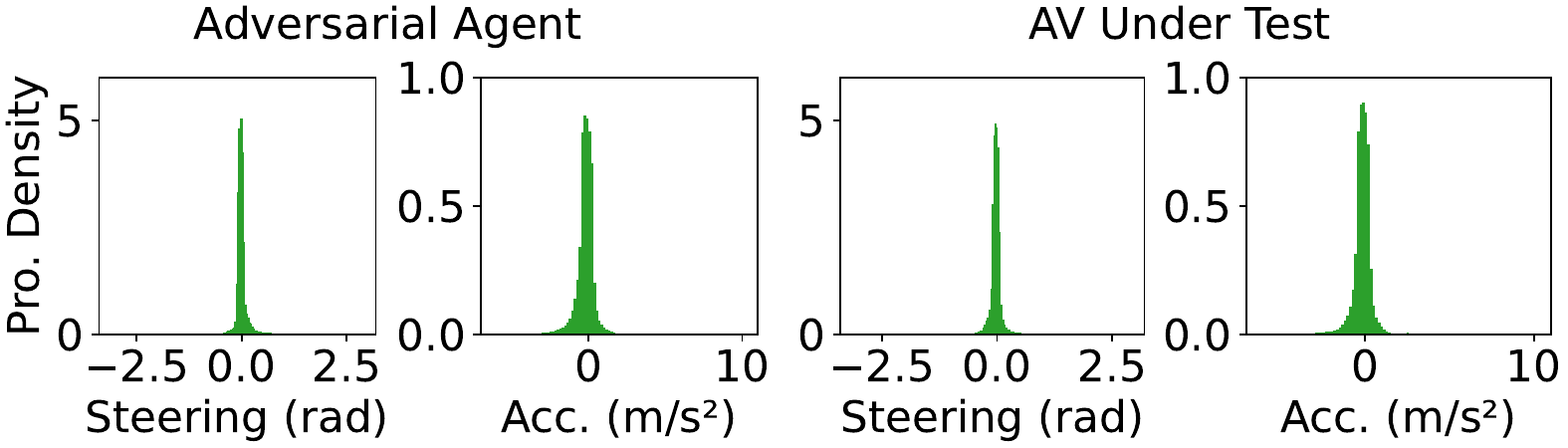}
    \caption{General driving scenarios.}
  \end{subfigure}
  \begin{subfigure}{\linewidth}
    \includegraphics[width=\linewidth]{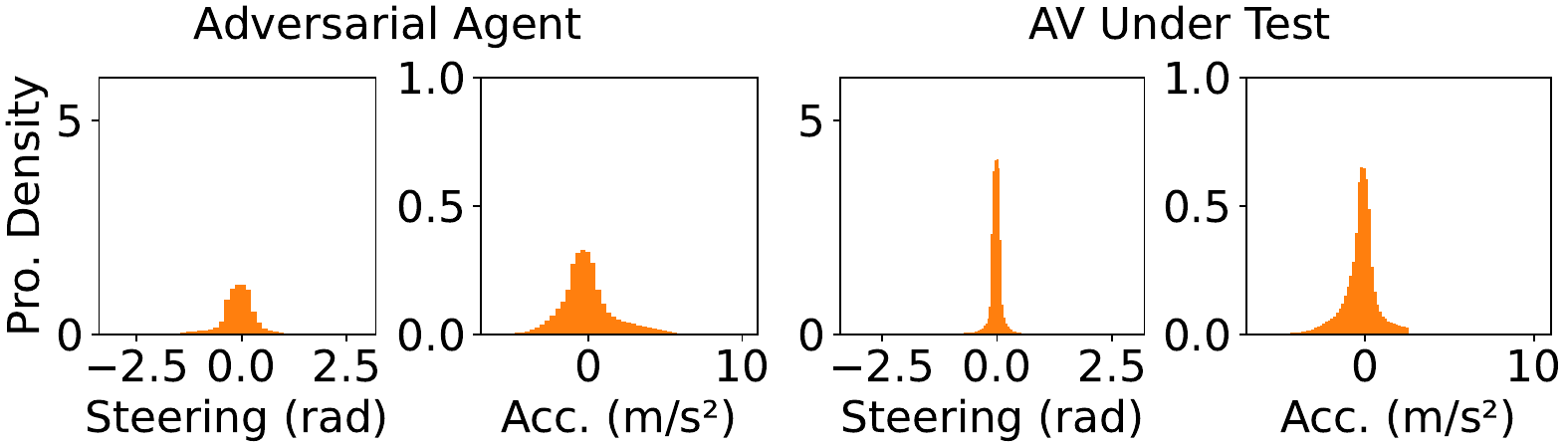}
    \caption{Baseline.}
  \end{subfigure}
  \begin{subfigure}{\linewidth}
    \includegraphics[width=\linewidth]{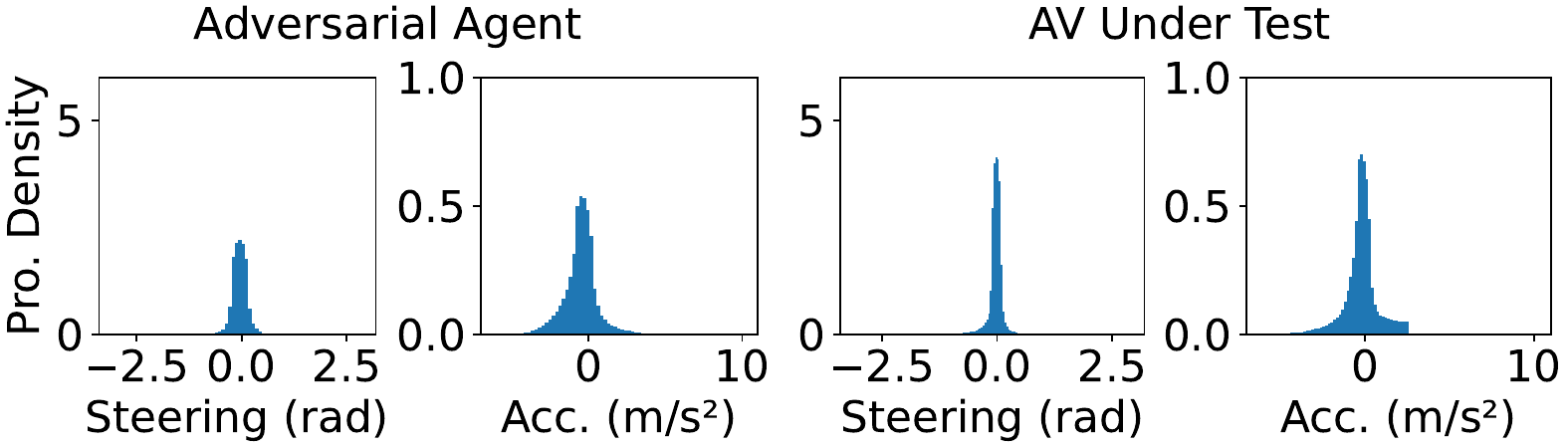}
    \caption{Proposed.}
  \end{subfigure}
  \caption{Statistical comparison among different solutions.}
  \label{fig:stat}
\end{figure}

\begin{table}
    \centering
    \caption{Comparisons of the Collision Rates and the Number of Lane Changes}
    \begin{tabular}{cccc} 
        \toprule
        \multirow{2}{*}[-.5ex]{{\bf Method}} & \multicolumn{2}{c}{{\bf Collision rate}} & \multirow{2}{*}[-.5ex]{{\bf \makecell{The No. of\\lane changes}}} \\ 
        \cmidrule(lr){2-3} 
        & {\bf AV under test} & {\bf Other vehicles} & \\
        \midrule
        Baseline & 62.84\% & 9.09\% & 2155 \\
        {\bf Proposed} & {\bf 42.17\%} & {\bf 5.79\%} & {\bf 2109} \\
        \bottomrule
    \end{tabular}
    \label{tab_3}
\end{table}

\begin{table}
\caption{Comparisons of the Acceleration and Steering.}
\label{tab_4}
\centering
\begin{tabular}{ccccc}
\toprule
& \multicolumn{2}{c}{{\bf Adversarial agent}} & \multicolumn{2}{c}{{\bf AV under test}} \\
\cmidrule(lr){2-3}\cmidrule(lr){4-5}
 & \makecell{Acceleration\\\unit{(m/s^{2})}} & \makecell{Steering\\\unit{(rad)}} & \makecell{Acceleration\\\unit{(m/s^{2})}} & \makecell{Steering\\\unit{(rad)}}\\
\midrule
Nor. & [-3.09, 2.08] & [-0.73, 0.72] & [-3.12, 2.05] & [-0.74, 0.72] \\
Bas. & [-6.69, 10.58] & [-3.38, 3.15] & [-5, 3] & [-1.05, 1.05] \\
{\bf Pro.} & {\bf [-4.94, 6.25]} & {\bf [-1.64, 1.58]} & {\bf [-5, 3]} & {\bf [-1.05, 1.05]} \\
\bottomrule
\end{tabular}
\end{table}

We present the results under the macroscopic traffic metrics in Table.~\ref{tab_3}. Consistent with our previous observation in Fig.~\ref{fig:reward_compare}, the baseline solution owns 20.67\% and 3.3\% more collisions for AV under test and other vehicles, respectively, and 45 times more lane changes in total. We then present the results under the microscopic traffic metrics in Table.~\ref{tab_4}. For the microscopic traffic metrics, the statistical results of acceleration and steering of the adversarial agent and the AV under test are shown in Table.~\ref{tab_4}. Note that in all settings, the AV under test is controlled by the surrogate model, and the ego vehicle is set as the adversarial agent, except for the normal one where the ego vehicle is controlled by the surrogate model as well. From the table, we can observe that the ranges of acceleration and steering from the baseline solution are much wider compared with the proposed one, meaning the baseline solution generates more adversarial scenarios that challenge the AV under test more often. However, given these kinds of wider maneuver intervals, the adversarial agent can behave unnaturally, e.g., with excessive acceleration, excessive deceleration, or excessive steering, resulting in T-shaped or opposite collisions as discussed in the Introduction. Instead of losing naturalness as the baseline solution, the solution we proposed narrows the interval of acceleration and steering by 35.21\% and 50.69\%, respectively, which maintains the naturalness while preserving the adversariality. To see these results more clearly, we draw the probability density distributions of the acceleration and steering in Fig.~\ref{fig:stat}, and this figure double proves the claim we have made, as the proposed solution performs closer to a natural driving scenario rather than the baseline solution that only concerns the adversariality.

\begin{figure*}
\centering
\includegraphics[width=\linewidth]{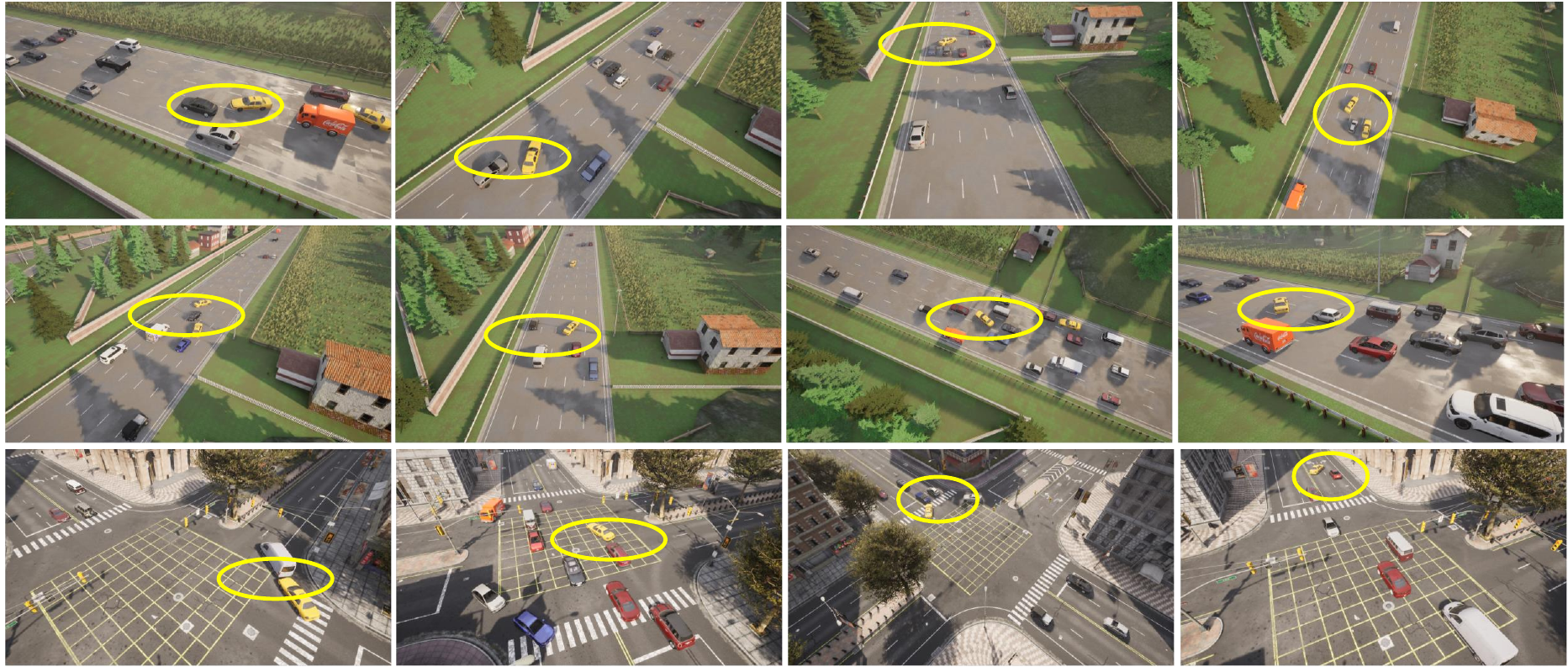}
\caption{Examples of the adversarial scenarios generated by our solution.}
\label{fig:demo}
\end{figure*}

\begin{figure}
\centering
\includegraphics[width=3 in]{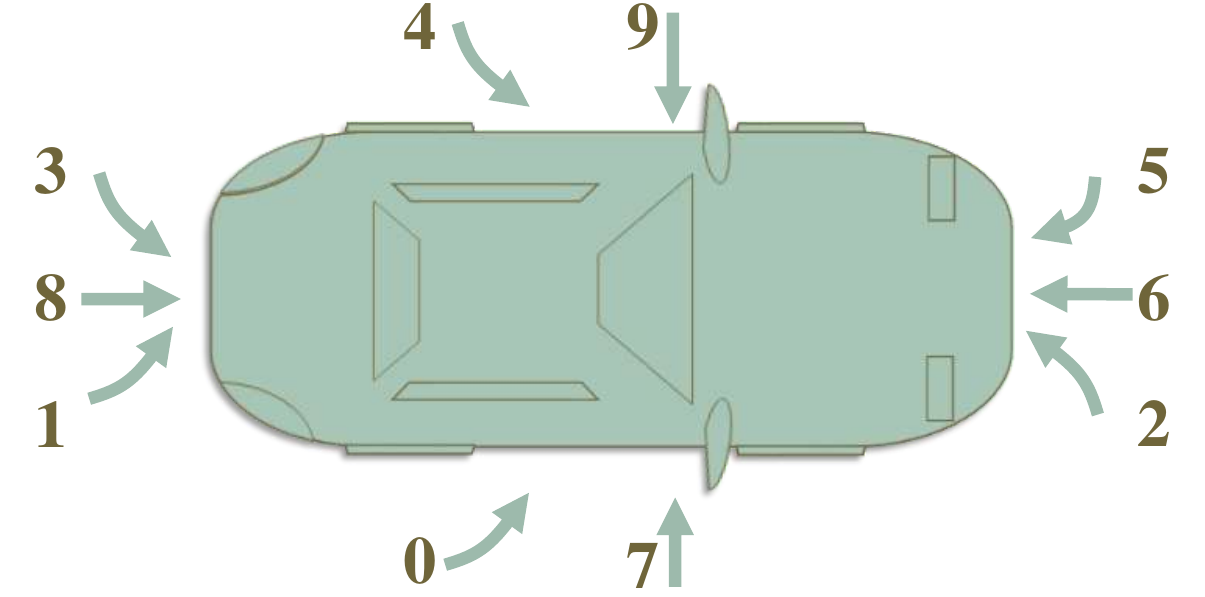}
\caption{Typical collision directions.}
\label{fig_9}
\end{figure}

\subsubsection{Clustering Analysis of Collision Types}

Having analyzed the results of the trained model and the key metrics, we now describe the underlying reasons for the superiority of our proposed solution in obtaining both naturalness and adversariality. To do this, we first define major collision types in general traffic, and then label the collisions in scenarios generated by the baseline model and our solution. Based on the labeled collisions, we can better understand the rationality behind our solution. In the meantime, by observing the collision type, we can verify whether we can guarantee diversity in the generated scenarios.

Specifically, given collisions often occur in a very short time, to capture the causes of collisions as precise as possible across different scenarios, critical features among all features have to be used, and after a thorough analysis, we decide to use the five features for this task, i.e., the relative lateral distance, relative longitudinal distance, relative lateral speed, relative longitudinal speed, and relative steering. In order to do that, we use the Principal Component Analysis~(PCA) dimensionality reduction technique~\cite{abdi2010principal} to reduce the dimension of the feature vectors from high dimensions, and only extract the wanted critical features. In doing this, we can use K-means clustering~\cite{alsabti1997efficient}, an effective unsupervised clustering technique in distinguishing data from different clusters, to label collision types according to these features.

\begin{figure}
\centering
\includegraphics[width=.9\linewidth]{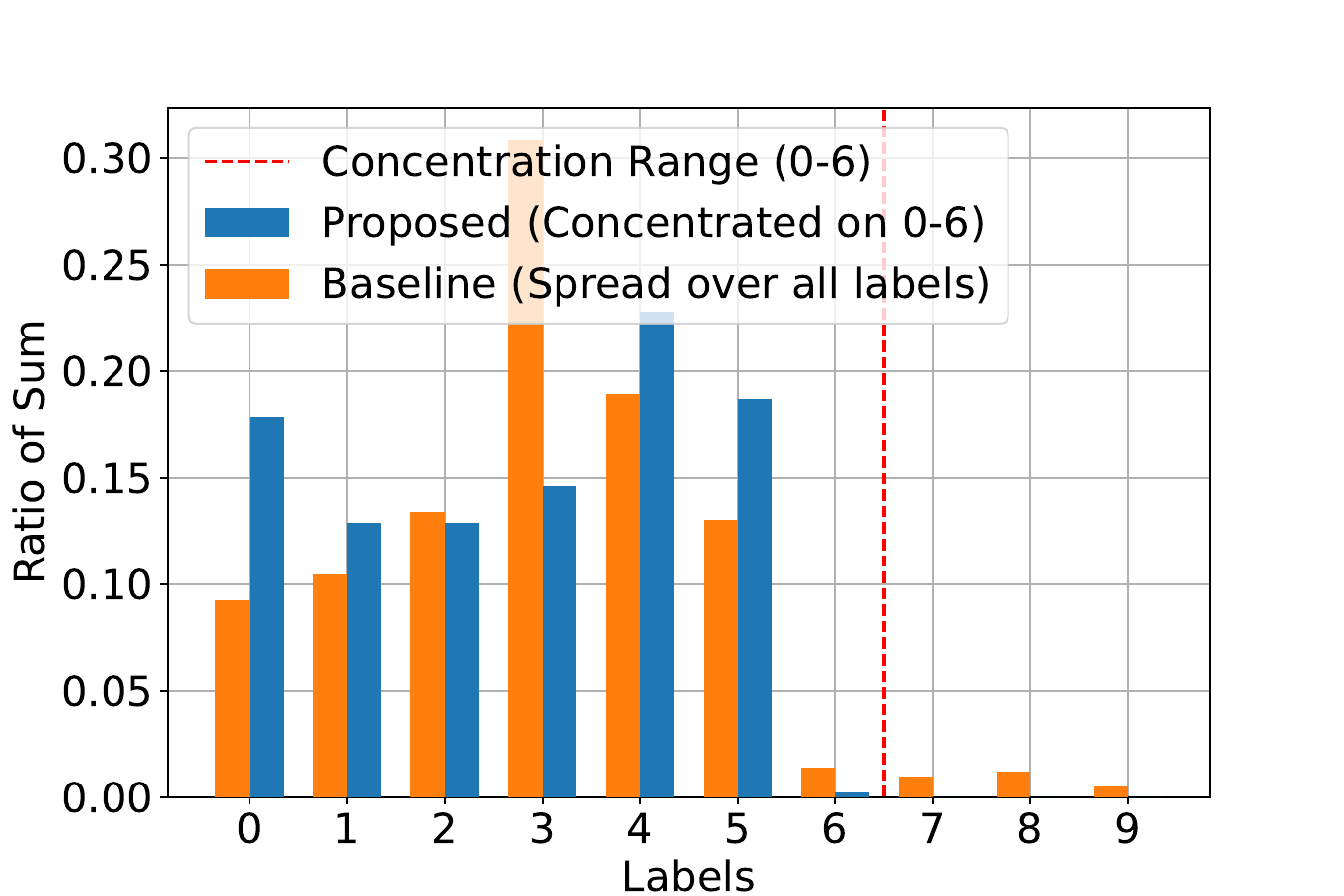}
\caption{Clustering results comparison.}
\label{fig:labels}
\end{figure}

Before labeling, we categorized 10 collision types as shown in Fig.~\ref{fig_9}, in which we give a number to each collision type from 0 to 9 and use an arrow to indicate the collision direction of a vehicle. Accordingly, collision label 7, 8, and 9 present counter-intuitive collisions, such as collisions from the rear at an extremely high speed, T-shaped collisions, and rear-to-rear collisions. In our experiment, we semantically labeled the collisions and plotted the ratios of each collision label in Fig.~\ref{fig:labels}. The collisions generated from our solution are all categorized as natural collisions. In contrast, the baseline solution generated a higher number of counter-intuitive collisions, specifically those labeled as 7, 8, and 9. This result proves our advantage in generating adversarial scenarios while persevering naturalness, as evidenced by avoiding counter-intuitive collisions. Furthermore, we offer more illustrative examples of adversarial scenarios generated by our solution in Fig.~\ref{fig:demo}, showcasing various scenarios on both highways and intersections.

\section{Conclusion and Future Work}
\label{conculsion}
By utilizing human driving priors and advanced reinforcement learning techniques, we have presented a natural adversarial test scenarios generation solution designed specifically for autonomous driving systems. Based on the observation that past work was overly concerned with the adversariality when generating safety-critical scenarios, resulting in many counter-intuitive scenarios which are both ineffective and misleading, our solution takes both adversariality and naturalness into account, and presents diverse adversarial scenarios that are practically close to real-world environments. The extensive experiments that have been conducted using real-world datasets demonstrate the superiority of our work over existing work.

Our method exhibits potential in various traffic situations, yet faces challenges in complex scenarios, lacking thorough consideration of human and environmental factors. Future work will aim at enhancing model sophistication to handle intricate traffic dynamics, incorporating novel behavioral models for realistic human interactions, and embedding environmental conditions for a holistic simulation. Additionally, efforts will be directed towards bolstering the efficiency and robustness of our solution through optimized algorithms using advanced multi-agent network models, and robustness analysis on diverse ODD elements, paving the way for real-world applicability and collaborative endeavors with traffic experts to broaden the scope and impact of our research.

\bibliographystyle{IEEEtran}
\bibliography{sample-base}
\vspace{-5cm}
\begin{IEEEbiography}[{\includegraphics[width=1in,height=1.25in,clip,keepaspectratio]{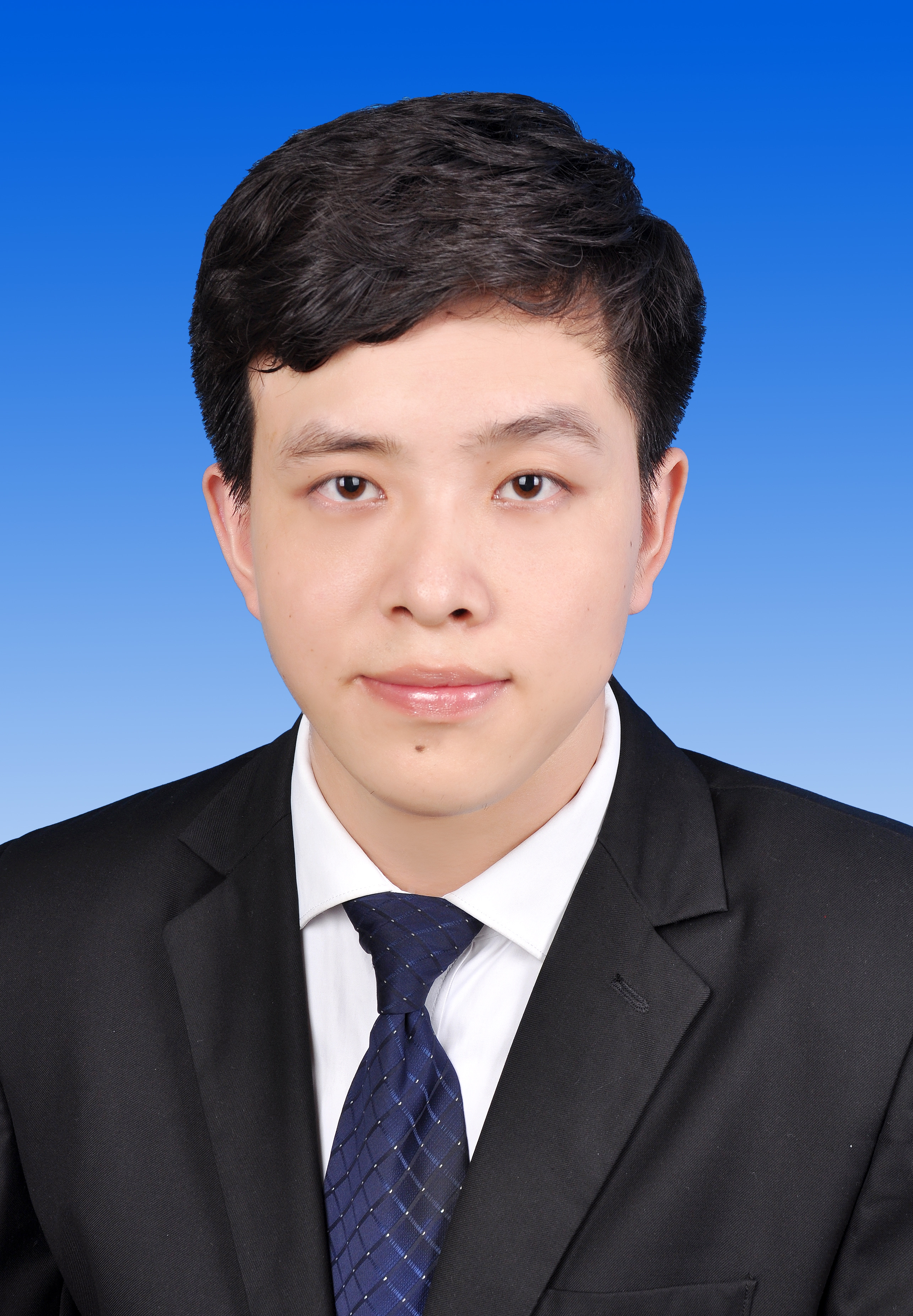}}]{Kunkun Hao}
received the M.S. degree in software engineering from Xi’an Jiaotong University, Xi’an, China, in 2019. He is currently working at the Research Center of Synkrotron, Inc., Xi’an, 710075, China. His research interests include reinforcement learning, vehicle safety of the intended functions, and traffic flow simulation.
\end{IEEEbiography}
\vskip -1.5\baselineskip plus -1fil
\begin{IEEEbiography}[{\includegraphics[width=1in,height=1.25in,clip,keepaspectratio]{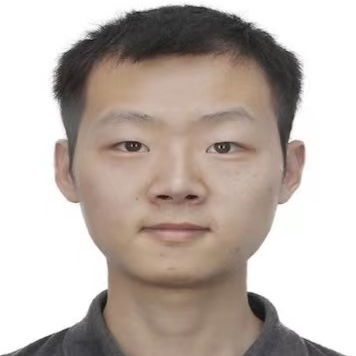}}]{Wen Cui}
(Member, IEEE) received the B.S. degree in computer science from Xi'an University of Science and Technology, Xi’an, China, in 2013, the M.S. degree in software engineering from Northwest University, Xi’an, China, in 2016, and the Ph.D. degree in electrical and computer engineering from the University of Victoria, Victoria, BC, Canada, in 2021. He is currently a postdoctoral researcher at the Institute for Interdisciplinary Information Core Technology, Xi’an, 710075, China, the Academy of Advanced Interdisciplinary Research, Xidian University, Xi’an, 710071, China, and also the Research Center of Synkrotron, Inc., Xi’an, 710075, China. His research interests include the design of autonomous driving systems, wireless communications, and vehicle-to-everything (V2X).
\end{IEEEbiography}
\vskip -1.5\baselineskip plus -1fil
\begin{IEEEbiography}[{\includegraphics[width=1in,height=1.25in,clip,keepaspectratio]{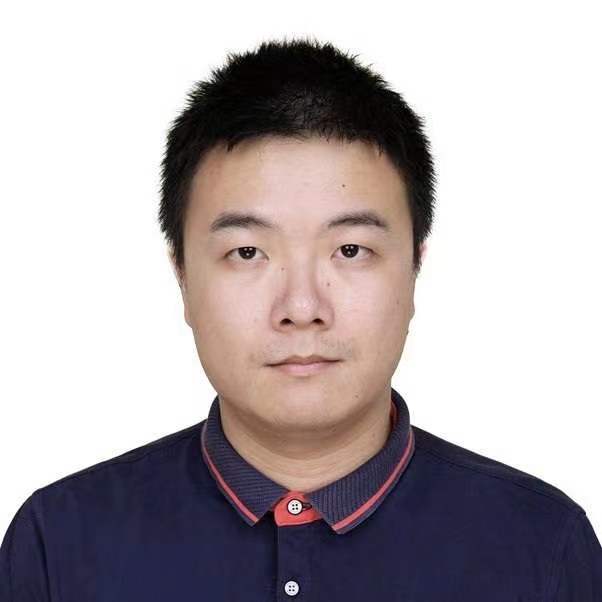}}]{Yonggang Luo}
is the director of AI Lab at Chongqing Changan Automobile. He obtained the Ph.D. degree from Purdue University in 2020, and the Bachelor Degree from Shandong University in 2014. His research interests span the areas of deep learning, especially including AI-based applications on vehicles.
\end{IEEEbiography}
\vskip -1.5\baselineskip plus -1fil
\begin{IEEEbiography}[{\includegraphics[width=1in,height=1.25in,clip,keepaspectratio]{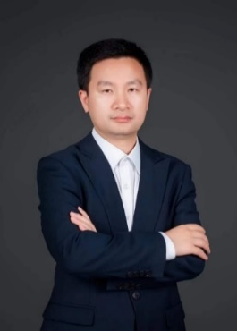}}]{Lecheng Xie}
serves as the Director of Chongqing Changan Automobile and as the Deputy Director of the AI \& SI Laboratory. He has led a project on Core Electronic Devices, High-end Generic Chips, and Basic Software under the Ministry of Industry and Information Technology. He has also participated in three national-level projects and one municipal-level project in Chongqing, earning the second prize in the Science and Technology Progress category within China's automobile industry.
\end{IEEEbiography}
\vskip -1.5\baselineskip plus -1fil
\begin{IEEEbiography}[{\includegraphics[width=1in,height=1.25in,clip,keepaspectratio]{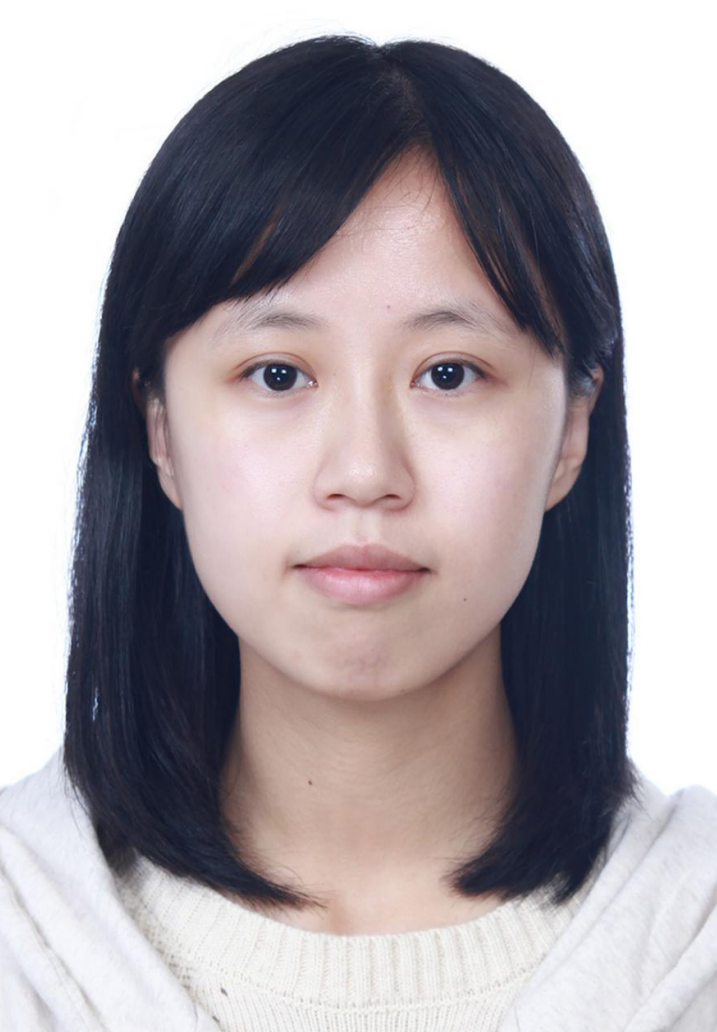}}]{Yuqiao Bai} received her Master's degree in 2021 from Leibniz University Hannover, Germany. She is currently working at the Research Center of Synkrotron, Inc., Xi’an, 710075, China, as an autonomous driving testing professional and contributes to various projects in the field. She holds a strong interest in the field of autonomous driving, especially in autonomous driving testing. 
\end{IEEEbiography}
\vskip -1.5\baselineskip plus -1fil
\begin{IEEEbiography}[{\includegraphics[width=1in,height=1.25in,clip,keepaspectratio]{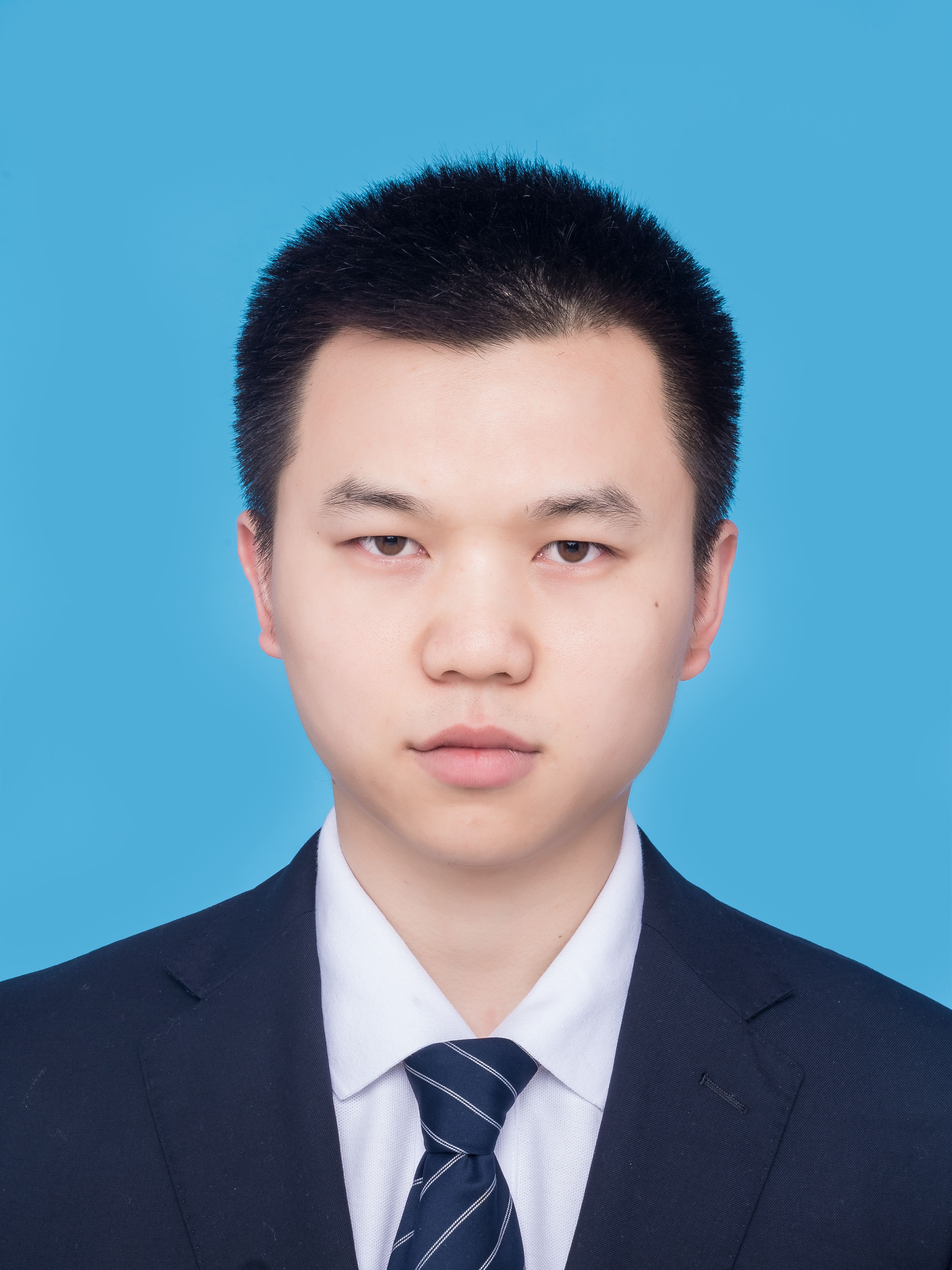}}]{Jucheng Yang}
is deputy chief engineer of decision cognition of AI Lab at Chongqing Changan Automobile. He graduated from Sichuan University in 2015. His main research fields are the cutting-edge algorithms related to automatic driving decision making, and the application of relevant algorithms in simulation and real vehicles.
\end{IEEEbiography}
\vskip -1.5\baselineskip plus -1fil
\begin{IEEEbiography}[{\includegraphics[width=1in,height=1.25in,clip,keepaspectratio]{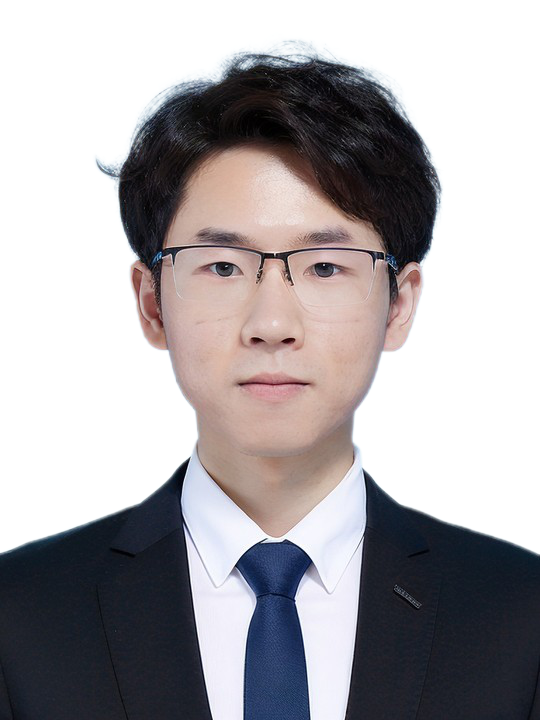}}]{Songyang Yan} received the B.S. degree in 2020 from the School of Software Engineering, Xi’an Jiaotong University, China. He is currently pursuing a Ph.D. in Xi’an Jiaotong University, China, specializing in Vehicle Safety of the Intended Functions (SOTIF) and adversarial scenario generation. He is one of the main developers of the Carla Leaderboard platform.
\end{IEEEbiography}
\vskip -1.5\baselineskip plus -1fil
\begin{IEEEbiography}[{\includegraphics[width=1in,height=1.25in,clip,keepaspectratio]{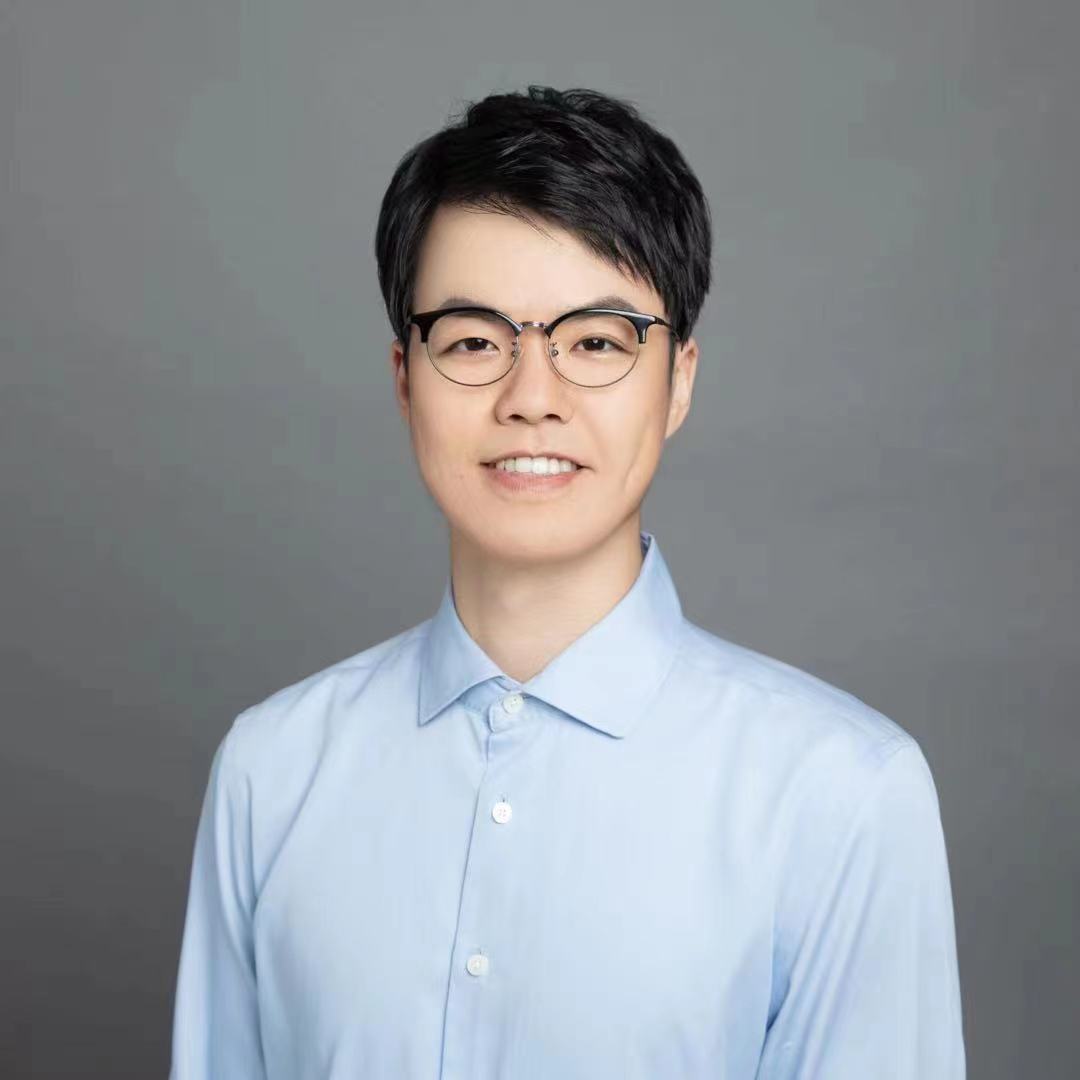}}]{Yuxi Pan}leads the R\&D team of Synkrotron Technologies, Inc. and works in the field of training and virtual validation of autonomous systems. He received Bachelor of Science degree from the University of Science and Technology of China in 2009, and PhD in Physics from the University of California, Los Angeles in 2015. He has served as a Technical Leader at Cisco Systems and Senior Applied Scientist at Uber, during which he was mainly engaged in research on artificial intelligence systems and their application in cloud infrastructure, information security and product experimentation.
\end{IEEEbiography}
\vskip -1.5\baselineskip plus -1fil
\begin{IEEEbiography}[{\includegraphics[width=1in,height=1.25in,clip,keepaspectratio]{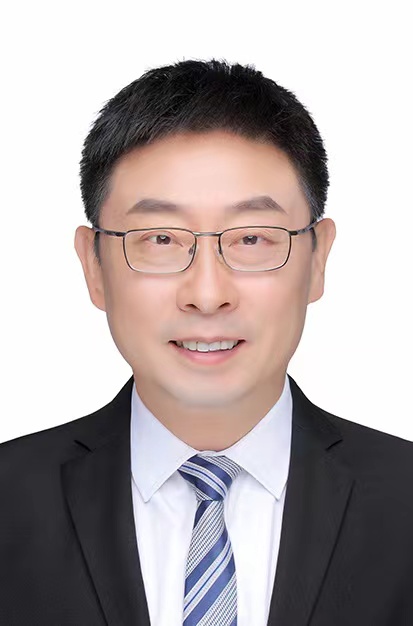}}]{Zijiang James Yang} (Senior Member, IEEE) holds a PhD from the University of Pennsylvania, a master's degree from Rice University, and a bachelor's degree from the University of Science and Technology of China.  He is the director of the Turing Interdisciplinary Information Science Research Center and a professor at Xi'an Jiaotong University. He is also the Founder and CEO of Synkrotron Inc. He served as a professor in the Department of Computer Science at Western Michigan University and a visiting professor in the Department of Electrical Engineering and Computer Science at the University of Michigan. Academic services include Chair of the IEEE Electric and Autonomous Vehicle Technical Committee; Vice Chair of the IEEE Autonomous Driving Standards Working Group; Chair of the 2021-2023 IEEE International Autonomous Driving Software Conference; Chair of the 2019 IEEE International Conference on Software Testing Verification and Validation; Chair of the Technical Executive Committee of the open source project Carsmos. Dr. Yang has published more than 100 papers and more than 20 Chinese and American patents. He received the ACM SIGSOFT Outstanding Paper Award, ACM TODAES Best Journal Paper Award, and Google Computer Science Engagement Award.
\end{IEEEbiography}
\vskip -1.5\baselineskip plus -1fil
\end{document}